\pdfoutput=1

\documentclass[11pt]{article}

\usepackage[final]{acl}
\usepackage{times}
\usepackage{latexsym}
\usepackage{colortbl}  
\usepackage[T1]{fontenc}

\usepackage[utf8]{inputenc}

\usepackage{microtype}
\usepackage{amsmath}
\usepackage{inconsolata}

\usepackage{graphicx}
\usepackage{booktabs}
\usepackage{multirow}
\usepackage{multicol}
\usepackage{array}
\usepackage{arydshln}

\title{\textbf{CoreEval}: Automatically Building Contamination-Resilient Datasets with Real-World Knowledge toward Reliable LLM Evaluation}

\author{
Jingqian Zhao$^{1\ast}$, 
Bingbing Wang$^{1}$\thanks{\quad The first two authors contribute equally to this work.}, 
Geng Tu$^{1}$,
Yice Zhang$^{1}$,
Qianlong Wang$^{1}$, \\
\bf Bin Liang$^{4\dagger}$,
Jing Li$^{5}$,
Ruifeng Xu$^{1, 2, 3}$\thanks{\quad Corresponding Author} \\
    $^{1}$ Harbin Institute of Technology, Shenzhen, China~
    $^{2}$ Peng Cheng Laboratory, Shenzhen, China \\
    $^{3}$ Guangdong Provincial Key Laboratory of Novel Security Intelligence Technologies \\
    $^{4}$ The Chinese University of Hong Kong, Hong Kong, China \\
    $^{5}$ The Hong Kong Polytechnic University, Hong Kong, China \\
    \texttt{\{zhaojingqian, bingbing.wang\}@stu.hit.edu.cn},
    ~\texttt{xuruifeng@hit.edu.cn}
}

\begin{document}
\maketitle
\begin{abstract}
Data contamination poses a significant challenge to the fairness of LLM evaluations in natural language processing tasks by inadvertently exposing models to test data during training.  
Current studies attempt to mitigate this issue by modifying existing datasets or generating new ones from freshly collected information. However, these methods fall short of ensuring contamination-resilient evaluation, as they fail to fully eliminate pre-existing knowledge from models or preserve the semantic complexity of the original datasets.
To address these limitations, we propose \textbf{CoreEval}, a \textbf{Co}ntamination-\textbf{re}silient \textbf{Eval}uation strategy for automatically updating data with real-world knowledge. This approach begins by extracting entity relationships from the original data and leveraging the GDELT database to retrieve relevant, up-to-date knowledge. The retrieved knowledge is then recontextualized and integrated with the original data, which is refined and restructured to ensure semantic coherence and enhanced task relevance. Ultimately, a robust data reflection mechanism is employed to iteratively verify and refine labels, ensuring consistency between the updated and original datasets.
Extensive experiments on updated datasets validate the robustness of CoreEval, demonstrating its effectiveness in mitigating performance overestimation caused by data contamination.
\end{abstract}

\begin{figure}[!t]
  \centering
  \includegraphics[width=\linewidth]{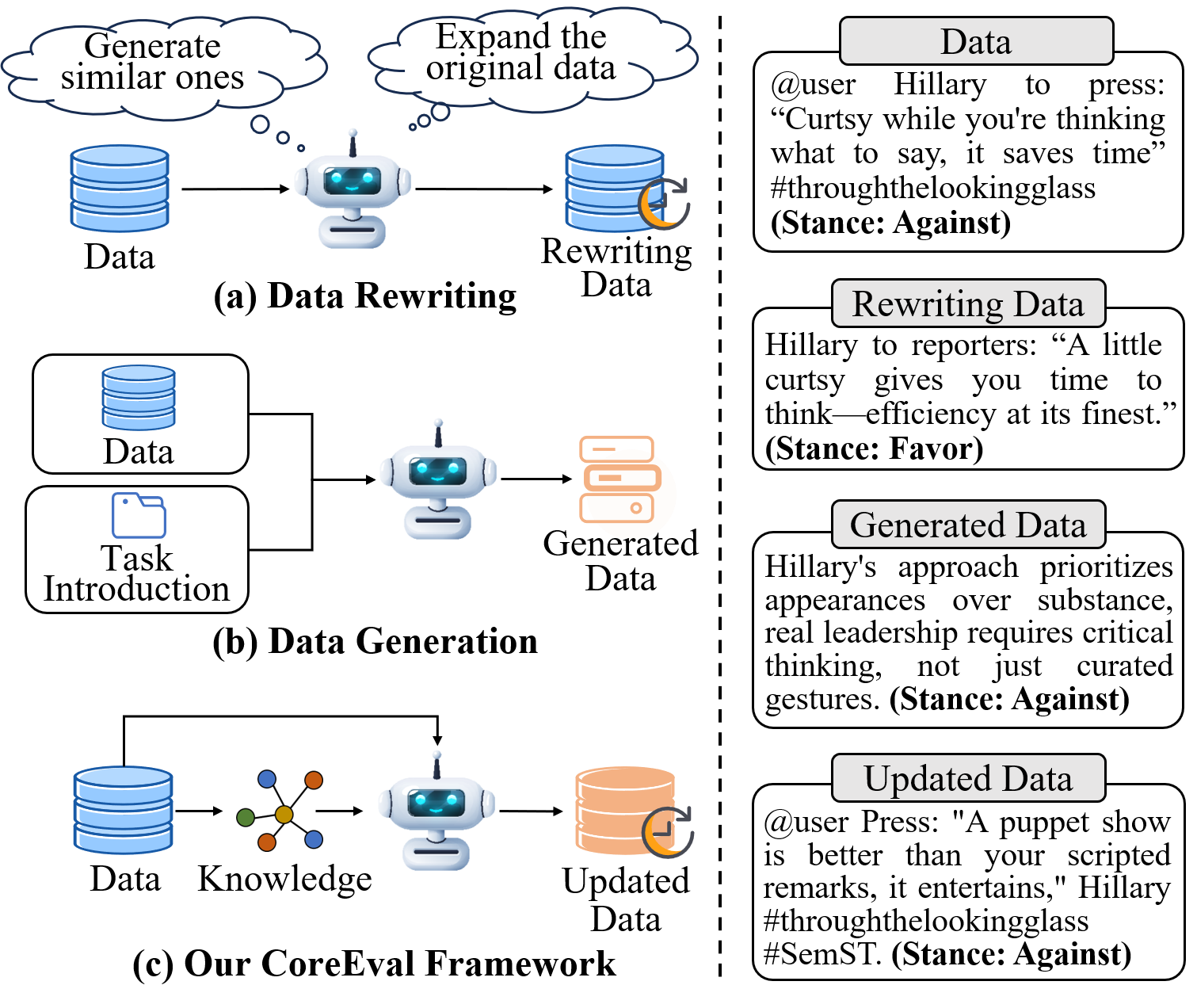}
  \caption{Different workflows for mitigating data contamination: (a) Data Rewriting, where LLMs modify existing data, potentially altering original labels; (b) Data Generation, where LLMs create new data from original data and task instructions, risking loss of semantic complexity; and (c) Our CoreEval Framework, where LLMs integrate external knowledge with original data for robust, semantically coherent, and label-consistent updates.}
  \label{f1}
  \vspace{-5pt}
\end{figure}
\section{Introduction}
In recent years, Large Language Models (LLMs) have demonstrated exceptional performance across a wide range of Natural Language Processing (NLP) tasks \cite{li2024ecomgpt,ma2024chain}. Publicly available datasets serve as standardized benchmarks for evaluating model performance, ensuring consistency and reproducibility in assessments. However, the static and public nature of these datasets poses a significant challenge: \textbf{data contamination}, where test data may inadvertently appear in the training sets of newer LLMs. This contamination can artificially inflate model performance, compromising the reliability of LLM evaluations \cite{banerjee2024vulnerability, li2024open}.

To mitigate data contamination, curating new datasets has become a widely adopted approach. Recently, researchers have explored automated dataset construction methods to reduce the time and labor costs associated with manual curation \cite{ying2024automating}. These approaches using LLMs can be broadly categorized into two types: data rewriting, which modifies existing data while preserving its original structure, and data generation, which leverages newly collected data to create task-specific datasets \cite{li2024latesteval, wu2024antileak}.

Despite their widespread adoption, these methods have significant limitations. 
As illustrated in Figure \ref{f1} (a), \textbf{data rewriting} employs prompt-based instructions to guide LLMs in modifying existing data. While this approach is straightforward, it often risks generating data with labels that deviate from the original annotations.  Additionally, the rewriting process may inadvertently introduce contaminated data, as models could rely on pre-existing information from their training corpus. 
On the other hand, \textbf{data generation}, which directly produces new datasets based on data and task introduction, shown in Figure \ref{f1} (b), fails to preserve the semantic richness and complexity of the original dataset, leading to information loss. These limitations undermine the reliability and effectiveness of existing approaches for contamination-resilient evaluation.

Therefore, this paper introduces \textbf{CoreEval}, a framework designed to mitigate data contamination and enable reliable, up-to-date LLM evaluation. As illustrated in Figure \ref{f1}, CoreEval goes beyond simple data rewriting and generation. Instead, it systematically integrates newly acquired knowledge, preserving data quality, enhancing robustness, and maintaining semantic richness while ensuring alignment with task objectives.
Specifically, CoreEval first extracts entity relationships from the original data and utilizes the Global Database of Events, Language, and Tone (GDELT) Project to retrieve up-to-date, real-world knowledge. This knowledge is then recontextualized with original data to refine and restructure the dataset, ensuring semantic coherence and alignment with task objectives.  
Finally, a rigorous data reflection mechanism enforces label consistency and preserves dataset integrity. We systematically evaluate CoreEval on multiple NLP datasets across different LLMs. Extensive experiments on these updated datasets validate the stability of our framework, demonstrating that CoreEval not only upholds high data quality but also effectively mitigates performance overestimation caused by data contamination.
The contributions of this paper can be summarized as follows:
\begin{itemize}
    \item We propose CoreEval, an automatic contamination-resilient evaluation strategy that integrates real-world knowledge to update datasets.
    \item We design a structured workflow inspired by cognitive learning theory to ensure reliable and timely LLM evaluation.
    \item Extensive experiments across multiple tasks and a series of LLMs demonstrate the effectiveness of CoreEval in mitigating data contamination.
\end{itemize}

\section{Related Works}
\subsection{Data Contamination}
Many datasets are widely used to evaluate models in NLP tasks like sentiment analysis \cite{saif2013evaluation, rogers2018rusentiment}, stance detection \cite{li2021p, glandt2021stance}, and emotion classification \cite{chen2017improving}. With LLMs, it is often assumed that a more advanced base model yields superior performance \cite{pathak2024comparative}. 
However, despite their critical role in benchmarking, the lack of transparency regarding the training data of these models makes it challenging for researchers to verify whether a given model has been contaminated by specific datasets.

Recent studies have explored data contamination in the evaluation of LLM. \citet{aiyappa2023can} analyzed ChatGPT’s stance detection, highlighting risks associated with its closed nature and updates. \citet{li2024open} reported contamination rates from 1\% to 45\% across six Question Answering (QA) benchmarks.
To tackle these challenges, researchers have explored methods for detecting contamination, revealing the limitations of string-matching techniques like n-gram overlap \citep{yang2023rethinking, jiang2024does, ippolito2023preventing}. Simple test variations, such as paraphrasing, can bypass these methods, allowing even a 13B model to overfit benchmarks and perform comparably to GPT-4. \citet{dekoninck2024evading} further emphasized these issues with the introduction of Evasive Augmentation Learning (EAL).

\begin{figure*}[!t]
  \centering
  \includegraphics[width=\linewidth]{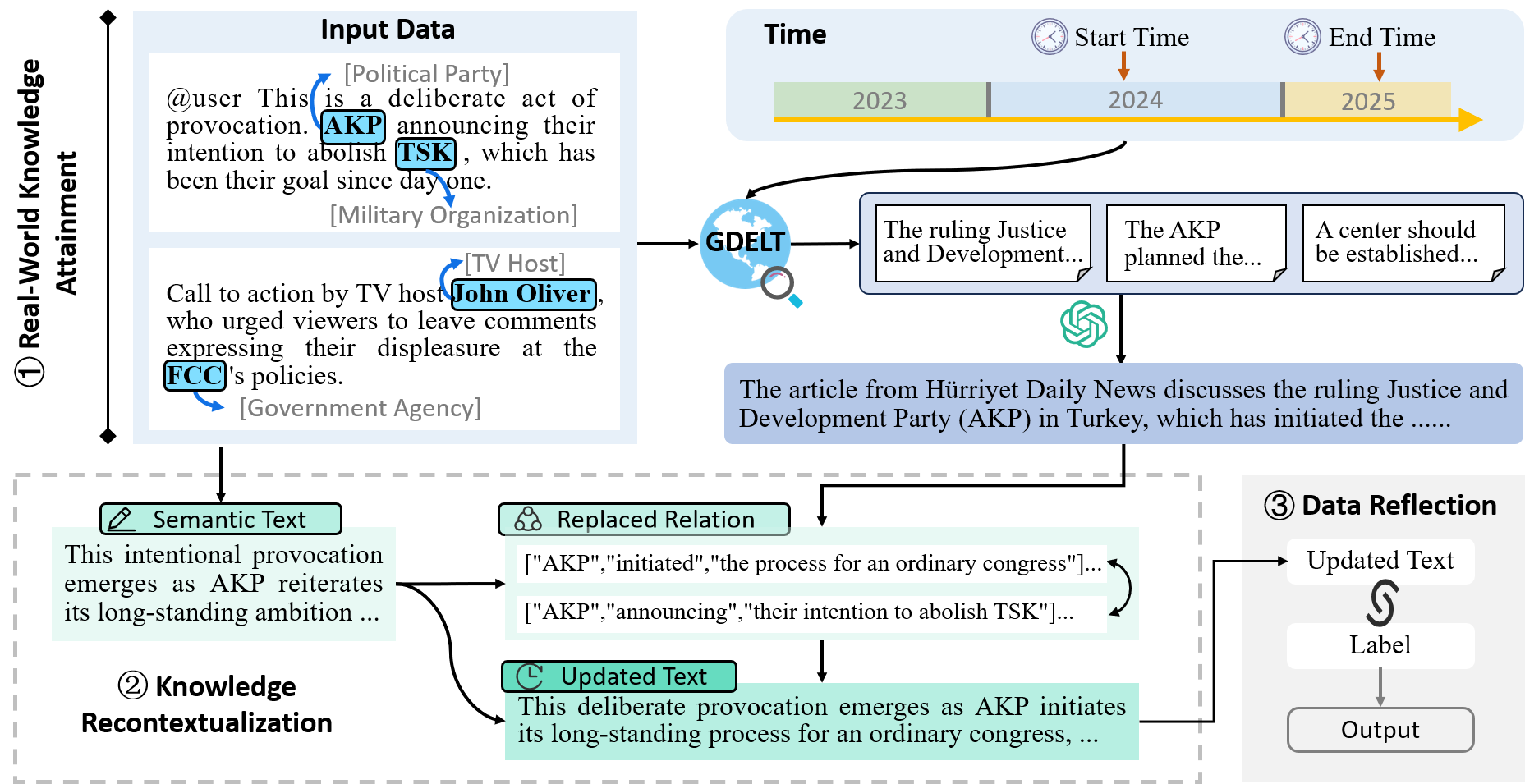}
  \caption{Overall flow of our CoreEval framework.}
  \label{f2}
\end{figure*}

\subsection{Contamination-Resilient Method}
To achieve contamination-resilient evaluation, updating datasets by collecting new data is an intuitive solution. However, due to the time-consuming and labor-intensive nature of this process, automatic update methods have emerged \cite{wu2024antileak}. These methods primarily fall into two categories: data rewriting and data generation.

Data rewriting modifies existing data to generate updated versions. \citet{ying2024automating} proposed two strategies: mimicking, which preserves style and context, ensuring consistency, and extending, which introduces varied difficulty to broaden the dataset’s cognitive scope.
Data generation relies on newly collected data to build task-specific datasets. LatestEval \cite{li2024latesteval} ensures integrity by using texts from recent sources, avoiding overlaps with pre-trained corpora. Similarly, LiveBench \cite{white2024livebench} creates novel datasets by extracting challenges from up-to-date sources like math competitions, arXiv papers, news articles, and transforming them into more challenging, contamination-free versions.
Despite their innovations, these methods have limitations. Data rewriting may produce inconsistent labels and introduce contamination from model biases, while data generation often fails to fully capture the semantic depth of the original dataset, leading to information loss. These challenges reduce the reliability and practicality of datasets for contamination-resilient evaluations.
Unlike these studies, CoreEval combines structured knowledge retrieval, semantic recontextualization, and iterative label verification to ensure dataset quality and robustness. By utilizing real-world updates and a reflection mechanism, CoreEval mitigates contamination while preserving semantic complexity.

\section{CoreEval Framework}
\subsection{Preliminary}
In this section, we introduce our novel CoreEval framework, inspired by Bruner’s cognitive theory, for constructing contamination-resilient datasets that integrate real-world knowledge. 
Building upon Bruner’s cognitive learning theory \cite{bruner2009process}, we assert that the essence of learning lies in the active formation of cognitive structures rather than the passive absorption of information. Learners actively construct their own knowledge systems by synthesizing newly acquired knowledge with their existing cognitive frameworks. Learning is conceptualized as involving three nearly simultaneous processes: the acquisition of information, the transformation of information, and its subsequent evaluation.
As shown in Figure~\ref{f2}, we organize these processes into three components to better align with LLM evaluation. \textbf{1) Real-World Knowledge Attainment} corresponds to information acquisition, collecting real-time knowledge from the GDELT database. \textbf{2) Knowledge Recontextualization} component handles information transformation, updating the dataset by incorporating new knowledge. \textbf{3) Data Reflection} component addresses the evaluation process by refining and assessing the data. This structure ensures that all learning processes are effectively integrated into a cohesive framework.

\subsection{Real-World Knowledge Attainment}
To incorporate real-world knowledge, we leverage GDELT \cite{leetaru2013gdelt}, a comprehensive CAMEO-coded database containing over 200 million geolocated events spanning global coverage from 1979 to the present. 
Given a dataset $\mathcal{D} = \{(d_1,y_1),(d_2,y_2),...,(d_n,y_n)\}$ consisting of $n$ samples, where each sample $d_i$ is paired with a corresponding label $y_i$ from the label set $\mathcal{Y}=\{y_1,y_2,...,y_n\}$. The knowledge extraction process begins by identifying relevant entities from the data using LLM $\mathcal{M}$, where the input $d_i$ acts as information cues for entity extraction.
\begin{equation}
    E_i \leftarrow \mathcal{M}(d_i)
\end{equation}
where $E_i=\{e_{i,1},e_{i,2},...,e_{i,{j_i}}\}$ and $j_i$ represents the set and number of entities extracted from $d_i$.
These extracted entities form the foundation for subsequent knowledge retrieval. To efficiently query large-scale data, we utilize Google BigQuery\footnote{\url{https://cloud.google.com/bigquery}} and the GDELT.
BigQuery enables fast, scalable processing of vast datasets like GDELT, while the API facilitates seamless real-time data retrieval.
A list of extracted entities is used to query GDELT databases $\mathcal{G}$ for data points within a specific time period to retrieve the most relevant and up-to-date knowledge.  Then we employ LLM to summarize the knowledge to obtain. The overall retrieval process can be formalized as:
\begin{equation}
\begin{split}
\mathcal{K}_i &\leftarrow \mathcal{G}(E_i,t_{\text{start}},t_{\text{end}}) \\
\mathcal{\hat K}_i &\leftarrow \mathcal{M}(\mathcal{K}_i)
\end{split}
\end{equation}
where $\mathcal{K}_i$ indicates the knowledge retrieved from the GDELT database. $\mathcal{\hat K}_i$ represents the knowledge after being summarized by the LLM.
$t_{\text{start}}$ and $t_{\text{end}}$ represent the start and end times for the query\footnote{We chose the release date of the latest open-source model as the starting point for retrieval to prevent overlap with the model's training data.}.

\subsection{Knowledge Recontextualization}
The knowledge recontextualization phase involves integrating new knowledge with existing cognitive structures, transforming it into a form suited for new tasks. During this phase, learners process and reorganize newly acquired knowledge to enhance both understanding and application. 
We begin by extracting relational triples from the original sentence $d_i$. These relational triples are represented as $T_i = \{\langle e_{i,j},r_{i,j},e^{'}_{i,j} \rangle \mid j=1,2,...,l_i \}$,
where $e_{i,j}$ and $e^{'}_{i,j}$ are entities, and $r_{i,j}$ denotes the relation between them. $l_i$ is the number of relational triples extracted from $d_i$. 
Next, using new knowledge $\mathcal{\hat K}_i$ and an LLM $\mathcal{M}$, 
we update the original triples $T_i$ by generating replacement triples $\hat T_i$.
The updated sentence $d^u_i$ is then derived by substituting the original triples with $\hat T_i$, as shown by:
\begin{equation}
\begin{split}
    \hat T_i &\leftarrow \mathcal{M}(T_i,\mathcal{\hat K}_i)\\
    d^u_i &\leftarrow f(d_i,\hat T_i)
\end{split}
\end{equation}
where $f$ is the replacement operation.

Furthermore, semantic rewriting is performed while preserving the $T_i$, resulting in: 
\begin{equation}
    d^s_i\leftarrow \mathcal{M}(d_i,T_i)
\end{equation}
We leverage the semantic style of $d^s_i$ combined with the label $y_i$ to construct a \textbf{semantic dataset $\mathcal{D}^s$}.

The updated text $\hat d_i$ adopts the semantic style of $d^s_i$, preserving its linguistic characteristics while incorporating the triples of $\hat T_i$. 
Additionally, to maintain classification coherence, the label of $\hat d_i$ is kept consistent with that of the original sentence  $d_i$. Formally, this process is represented as:
\begin{equation}
\hat d_i \leftarrow \mathcal{M}(d_i,d^u_i,\hat T_i,d^s_i)    
\end{equation}

The \textbf{updated dataset} $\mathcal{\hat D}$ is then formed by combining $\hat d_i$ with the corresponding label.
This process ensures the systematic integration of new knowledge while maintaining the coherence and adaptability of the transformed content.

\subsection{Data Reflection}
To evaluate the quality of the generated text, we design an agent to reflect and perform evaluations. This evaluation process employs prompting \cite{wei2022chain} to facilitate step-by-step reasoning. 
The assessment focuses on two key criteria:

\textbf{Incorrect Information}: Evaluating whether the generated text accurately reflects the facts derived from the provided knowledge. Any discrepancies or inconsistencies are flagged for re-generation.

\textbf{Label Alignment}: Measuring the degree of alignment between the generated text and the corresponding ground truth label, ensuring consistency and relevance to the intended output.

The prompting allows the agent to iteratively reflect on these criteria, providing a rationale for its evaluation. Based on this reflection, the agent determines whether the text required to be regenerated to improve accuracy or alignment. Detailed prompts can be found in Appendix~\ref{appendix-prompt_framework}.

\subsection{Apply to Existing Datasets}
We selected five representative Natural Language Understanding (NLU) tasks from the TweetEval Benchmark~\cite{barbieri2020tweeteval} and GLUE Benchmark~\cite{wang2018glue}, including Emotion Recognition~\cite{mohammad2018semeval}, Irony Detection~\cite{van2018semeval}, Stance Detection~\cite{mohammad2016semeval}, Microsoft Research Paraphrase Corpus (MRPC)~\cite{dolan2005automatically}, and  Recognizing Textual Entailment (RTE)~\cite{wang2018glue}, to apply our method for automatic updating and evaluation. Table~\ref{tab-dataset_info} presents the statistical characteristics of these datasets.
Notably, for the MRPC and RTE datasets, we refine the provided sentence pairs during the data reflection phase and ensure the supervision of label accuracy for improved consistency and correctness.

\begin{table}[!ht]
\setlength{\tabcolsep}{4.0pt}
\renewcommand{\arraystretch}{1.1}
\small
\centering
\begin{tabular}{lrrc}
\toprule
\textbf{Dataset} & 
\multicolumn{1}{c}{\textbf{Train}} &
\multicolumn{1}{c}{\textbf{Test}} &
\textbf{Label Space}  \\ \midrule
Emotion & 3,257      & 1,421     &  joy, optimism, sadness, anger           \\
Irony   & 2,862      & 784      &    irony, not irony         \\
Stance  & 2,620      & 1,249     &   favor, against, neutral        \\
MRPC    & 4,076      & 1,587     &   equivalent, not equivalent         \\
RTE     & 2,490      & 277      &    entailment, not entailment         \\ 
\bottomrule
\end{tabular}
\caption{Statistical overview of the five datasets, detailing training and test set sizes along with their corresponding task labels.}
\label{tab-dataset_info}
\end{table}

\subsection{Human Verification on Data Quality}
To ensure the reliability of our proposed strategy, we conduct a comprehensive human evaluation with five experienced computational linguistics researchers. All evaluators underwent prior training to ensure consistency in their assessments.
The evaluators analyze 50 randomly selected samples based on four key criteria:
\textbf{Fluency}, 
\textbf{Coherence}, 
\textbf{Factuality}, and 
\textbf{Accuracy}.
Following the approach of \citet{ying2024automating}, Fluency and Coherence are rated on a 3-point scale: 2 (Good), 1 (Acceptable), and 0 (Unsatisfactory). Factuality and Accuracy are rated as 1 (Yes) or 0 (No).
Detailed evaluation guidelines can be found in Appendix~\ref{appendix-guideline}.

To assess inter-annotator agreement, we use Fleiss' Kappa Statistic \cite{fleiss1971measuring}. 
As shown in Table \ref{kappa}, the results demonstrates that our method generates high-quality data through proper demonstration and structured workflow. 
Moreover, the values of $\kappa$ falling within the range 0.70 < $\kappa$ < 0.85 indicate substantial agreement among annotators.

\begin{table}[!h]
\centering
\setlength{\tabcolsep}{1mm}
\resizebox{\linewidth}{!}{
\begin{tabular}{lccccc}
\toprule
 \textbf{Dataset}&	\textbf{Fluency}&\textbf{Coherence}&\textbf{Factuality}&\textbf{Accuracy}&\textbf{$\kappa$} \\ \midrule
Emotion&2.99	&2.55&	0.98	&0.94	&0.73 \\
Irony &2.97&	2.74	&0.99&	0.97&	0.78\\
Stance &2.99	&2.56	&0.98	&0.96&	0.73\\
MRPC&2.98&	2.92&	0.98	&0.96&	0.86
\\
RTE&2.99	&2.86&	0.96	&0.96&	0.80
\\
\bottomrule
\end{tabular}
}
\caption{The statistics of the updated datasets are presented. $\kappa$ denotes Fleiss’ Kappa \cite{fleiss1971measuring}.}
\label{kappa}
\end{table}

\section{Experiment}
This section first presents the experimental setups, including model configurations and metrics. We then address the following questions to assess the effectiveness of our CoreEval:
\textbf{Q1:} How does LLM performance change across different tasks after data updates? 
\textbf{Q2:} Does CoreEval outperform existing methods in resisting data contamination? 
\textbf{Q3:} How does the dataset perform under different contamination proportions and types?

\subsection{Experiment Setup}
\textbf{Large Language Models}. 
For our experimental investigation, we curated a diverse set of language models comprising eight widely-adopted \textbf{open-source LLMs}: Llama3-8B~\cite{dubey2024llama}, Llama2-13B~\cite{touvron2023llama}, Ministral-8B~\cite{ministraux}, 
Mistral-NeMo-12B~\cite{mistralnemo} (abbreviated as Mistral-12B), Yi1.5-6B~\cite{young2024yi}, Yi1.5-9B~\cite{young2024yi}, Qwen2.5-7B~\cite{qwen2.5}, and Qwen2.5-14B~\cite{qwen2.5}\footnote{For all aforementioned open-source models, we utilized instruction-tuned versions of the model weights.}.
The experimental evaluation also included three prominent \textbf{proprietary LLMs}: ChatGPT, Gemini1.5, and Claude3.5\footnote{In our experiments, we utilized the following model versions: gpt-3.5-turbo-0125 for ChatGPT, gemini-1.5-flash for Gemini1.5, and claude-3-5-haiku-20241022 for Claude3.5.}.

\textbf{Evaluation Metrics}. 
Inspired by ~\citet{opitz2024closer}, we adopted the macro F1-score as the unified evaluation metric across all tasks to ensure consistency in performance assessment.
Following~\citet{ying2024automating}, we evaluate the model's performance $P$ using the macro F1-score and subsequently employ performance gain as a metric to assess its resilience to data contamination. 
This metric quantifies the improvement from test set fine-tuning, with a smaller boost indicating greater resistance to contamination.
In the contamination test experiment, we implement two simulation settings. The first involves training solely on the test set and measuring the performance gain $\delta_1=P_{test}-P_{zero}$ against zero-shot performance 
where $P_{test}$ denotes performance after fine-tuning on the test set only, and $P_{zero}$ represents the zero-shot performance.
The second setting incorporates both training and test sets, comparing the performance gain $\delta_2=P_{train + test}-P_{train}$.
where $P_{train}$ indicates performance after fine-tuning on the training set alone, and $P_{train + test}$ represents performance after fine-tuning with both training and test sets.
Detailed information about metric $\delta$ can be found in Appendix~\ref{appendix-resistance_indicators}.

\subsection{Performance Test (Q1)}
\label{sec-performance_test}

\begin{figure*}[!t]
  \centering
  \includegraphics[width=\linewidth]{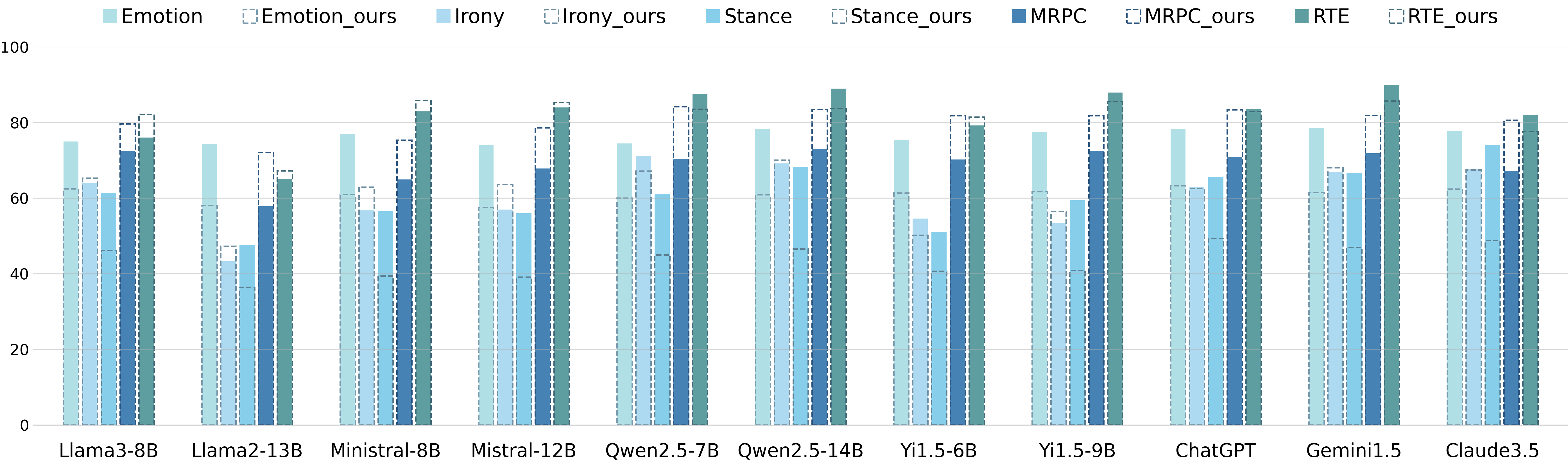}
  \caption{Performance (\%) of the eleven involved LLMs (zero-shot) on the original and our updated datasets. We employ various prompt templates and use their average as the final result. Refer to Appendix~\ref{appendix-performance} for further details.}
  \label{fig-f1score}
\end{figure*}

We first evaluate the zero-shot performance of LLMs on both the original and our updated datasets, using zero-shot evaluation as a standard configuration for assessing LLMs capabilities. We analyze how LLMs performance varies across different tasks after data updates.
Refer to Appendix~\ref{appendix-configuration_inference} for the inference configurations.
To mitigate prompt bias, we average results across multiple prompt templates, with 
detailed prompts provided in Appendix~\ref{appendix-prompt_experiment}.

The experimental results, illustrated in Figure~\ref{fig-f1score}, reveal the following: 
1) While \textbf{proprietary models generally outperform most open-source models, the Qwen2.5 series achieves comparable or even superior performance} among open-source models. 
2) \textbf{Emotion recognition and stance detection tasks substantially decline in performance on our updated dataset relative to the original one}. This decline can be attributed to two factors. 
First, these tasks may already be contaminated in existing LLMs, leading to decreased performance on our updated dataset, which aligns with prior studies~\cite{aiyappa2023can, sainz2024data}. Second, emotion and stance tasks inherently involve more subjective interpretations and contextual nuances, requiring an understanding of complex, evolving social and cultural contexts.
The injection of new knowledge can alter textual patterns, including time-dependent emotional and stance expressions, thereby affecting LLM judgments. This underscores the importance of timely LLM iterations.
3) Proprietary models exhibit a more significant performance drop of 5.42\%, compared to 3.62\% for open-source models, suggesting that \textbf{proprietary models may suffer from more severe data contamination}. The lack of transparency in their training data and model parameters makes detecting and mitigating data contamination in proprietary systems a critical challenge.

\subsection{Contamination Test (Q2)}
\label{sec-contamination_test}

\begin{table*}[t]
\small
\centering
\setlength{\tabcolsep}{3.5pt}
\renewcommand{\arraystretch}{1.1}
\begin{tabular}{llccccccccccccccccccc}
\hline
\multirow{2}{*}{}       & \multirow{2}{*}{} & & \multicolumn{2}{c}{Emotion}   &  & \multicolumn{2}{c}{Irony}      &  & \multicolumn{2}{c}{Stance}     &  & \multicolumn{2}{c}{MRPC}        &  & \multicolumn{2}{c}{RTE}         &  & \multicolumn{2}{c}{AVG}        \\ \cline{4-5} \cline{7-8} \cline{10-11} \cline{13-14} \cline{16-17} \cline{19-20} 
                             &  &                       & $\mathit{\delta_1} \downarrow$        & $\mathit{\delta_2} \downarrow$        &  & $\mathit{\delta_1} \downarrow$         & $\mathit{\delta_2} \downarrow$        &  & $\mathit{\delta_1} \downarrow$         & $\mathit{\delta_2} \downarrow$        &  & $\mathit{\delta_1} \downarrow$         & $\mathit{\delta_2} \downarrow$         &  & $\mathit{\delta_1} \downarrow$         & $\mathit{\delta_2} \downarrow$         &  & $\mathit{\delta_1} \downarrow$         & $\mathit{\delta_2} \downarrow$        \\ \hline
\multirow{3}{*}{Llama3-8B}  & \textit{orig} &                 & 9.37          & 4.47          &  & 30.09          & 7.07          &  & 23.41          & 6.80          &  & 10.98          & 7.05           &  & 20.88          & 8.79           &  & 18.95          & 6.84          \\
                             & \textit{semt} &               & 4.86          & 1.34          &  & 9.66           & 3.05          &  & 20.00          & 3.14          &  & 6.37           & 2.78           &  & 12.20          & \cellcolor[HTML]{EFEFEF}\textbf{0.12}          &  & 10.62          & 2.09          \\
                             & \textit{ours} &                   & \cellcolor[HTML]{EFEFEF}\textbf{3.27} & \cellcolor[HTML]{EFEFEF}\textbf{1.33} & \cellcolor[HTML]{EFEFEF} & \cellcolor[HTML]{EFEFEF}\textbf{2.00}  & \cellcolor[HTML]{EFEFEF}\textbf{1.89} & \cellcolor[HTML]{EFEFEF} & \cellcolor[HTML]{EFEFEF}\textbf{11.66}  & \cellcolor[HTML]{EFEFEF}\textbf{2.57} & \cellcolor[HTML]{EFEFEF} & \cellcolor[HTML]{EFEFEF}\textbf{0.75}  & \cellcolor[HTML]{EFEFEF}\textbf{0.53}  & \cellcolor[HTML]{EFEFEF} & \cellcolor[HTML]{EFEFEF}\textbf{4.21}  & 0.13  & \cellcolor[HTML]{EFEFEF} & \cellcolor[HTML]{EFEFEF}\textbf{4.38}  & \cellcolor[HTML]{EFEFEF}\textbf{1.29} \\ \hline
\multirow{3}{*}{Llama2-13B}   & \textit{orig} &                 & 11.83         & 4.55          &  & 52.46          & 7.97          &  & 38.26          & 6.98          &  & 26.98          & 6.83           &  & 26.24          & 9.02           &  & 31.16          & 7.07          \\
                             & \textit{semt} &               & 7.69          & 1.60          &  & 23.15          & 2.42          &  & 32.70          & 3.26          &  & 18.44          & 2.50           &  & 22.63          & 2.15           &  & 20.92          & 2.38          \\
                             & \textit{ours} &                   & \cellcolor[HTML]{EFEFEF}\textbf{7.41} & \cellcolor[HTML]{EFEFEF}\textbf{0.57} &\cellcolor[HTML]{EFEFEF}  & \cellcolor[HTML]{EFEFEF}\textbf{18.23} & \cellcolor[HTML]{EFEFEF}\textbf{1.12} & \cellcolor[HTML]{EFEFEF} & \cellcolor[HTML]{EFEFEF}\textbf{24.50} & \cellcolor[HTML]{EFEFEF}\textbf{2.56} &  \cellcolor[HTML]{EFEFEF}& \cellcolor[HTML]{EFEFEF}\textbf{10.09} & \cellcolor[HTML]{EFEFEF}\textbf{-0.31}  & \cellcolor[HTML]{EFEFEF} & \cellcolor[HTML]{EFEFEF}\textbf{21.12} & \cellcolor[HTML]{EFEFEF}\textbf{1.10}  & \cellcolor[HTML]{EFEFEF} & \cellcolor[HTML]{EFEFEF}\textbf{16.27} & \cellcolor[HTML]{EFEFEF}\textbf{1.01} \\ \hline
\multirow{3}{*}{Ministral-8B} & \textit{orig} &                 & 12.65         & 6.85          &  & 39.66          & 7.58          &  & 30.80          & 8.51          &  & 25.64          & 8.91           &  & 17.08          & 6.64           &  & 25.17          & 7.70          \\
                             & \textit{semt} &               & 6.77          & 1.58          &  & 10.53          & 1.98          &  & 28.47          & 3.38          &  & 11.94           & 2.84           &  & 6.50          & -0.72           &  & 12.84          & 1.81          \\
                             & \textit{ours} &                   & \cellcolor[HTML]{EFEFEF}\textbf{4.41} & \cellcolor[HTML]{EFEFEF}\textbf{0.15} &\cellcolor[HTML]{EFEFEF}  & \cellcolor[HTML]{EFEFEF}\textbf{2.54}  & \cellcolor[HTML]{EFEFEF}\textbf{0.58} & \cellcolor[HTML]{EFEFEF} & \cellcolor[HTML]{EFEFEF}\textbf{20.97} & \cellcolor[HTML]{EFEFEF}\textbf{2.36} & \cellcolor[HTML]{EFEFEF} & \cellcolor[HTML]{EFEFEF}\textbf{3.86}  & \cellcolor[HTML]{EFEFEF}\textbf{0.32}  & \cellcolor[HTML]{EFEFEF} & \cellcolor[HTML]{EFEFEF}\textbf{4.03} & \cellcolor[HTML]{EFEFEF}\textbf{-1.46} & \cellcolor[HTML]{EFEFEF} & \cellcolor[HTML]{EFEFEF}\textbf{7.16}  & \cellcolor[HTML]{EFEFEF}\textbf{0.39} \\ \hline
\multirow{3}{*}{Mistral-12B}  & \textit{orig} &                 & 17.41         & 7.59          &  & 40.43          & 10.59         &  & 34.69          & 9.35          &  & 26.51          & 9.30           &  & 13.43          & 7.49           &  & 26.50          & 8.86          \\
                             & \textit{semt} &               & 10.83         & \cellcolor[HTML]{EFEFEF}\textbf{1.44} &  & 8.46           & 4.27          &  & 30.49          & 3.69          &  & 10.54           & 2.28           &  & 2.74          & 0.11  &  & 12.61          & 2.36          \\
                             & \textit{ours} &                   & \cellcolor[HTML]{EFEFEF}\textbf{7.64} & 1.54          &  \cellcolor[HTML]{EFEFEF}& \cellcolor[HTML]{EFEFEF}\textbf{2.92}  & \cellcolor[HTML]{EFEFEF}\textbf{3.40} &\cellcolor[HTML]{EFEFEF}  & \cellcolor[HTML]{EFEFEF}\textbf{23.35} & \cellcolor[HTML]{EFEFEF}\textbf{3.19} & \cellcolor[HTML]{EFEFEF} & \cellcolor[HTML]{EFEFEF}\textbf{0.61}  & \cellcolor[HTML]{EFEFEF}\textbf{0.43}  & \cellcolor[HTML]{EFEFEF} & \cellcolor[HTML]{EFEFEF}\textbf{1.45}  & \cellcolor[HTML]{EFEFEF}\textbf{-0.62}           & \cellcolor[HTML]{EFEFEF} & \cellcolor[HTML]{EFEFEF}\textbf{7.19}  & \cellcolor[HTML]{EFEFEF}\textbf{1.59} \\ \hline
\multirow{3}{*}{Yi1.5-6B}    & \textit{orig} &                 & 11.42         & 4.65          &  & 39.78          & 8.45          &  & 31.75          & 8.64          &  & 14.96          & 7.71           &  & 19.64          & 8.80           &  & 23.51          & 7.69          \\
                             & \textit{semt} &               & 4.76          & \cellcolor[HTML]{EFEFEF}\textbf{0.60}          &  & 20.47          & 2.62          &  & 24.70          & 2.46          &  & 6.92           & 1.39           &  & 11.16          & \cellcolor[HTML]{EFEFEF}\textbf{0.36}  &  & 13.60          & 1.49          \\
                             & \textit{ours} &                   & \cellcolor[HTML]{EFEFEF}\textbf{3.50} & 0.84 & \cellcolor[HTML]{EFEFEF} & \cellcolor[HTML]{EFEFEF}\textbf{16.35} & \cellcolor[HTML]{EFEFEF}\textbf{0.75} & \cellcolor[HTML]{EFEFEF} & \cellcolor[HTML]{EFEFEF}\textbf{18.79} & \cellcolor[HTML]{EFEFEF}\textbf{2.45} & \cellcolor[HTML]{EFEFEF} & \cellcolor[HTML]{EFEFEF}\textbf{-1.00}  & \cellcolor[HTML]{EFEFEF}\textbf{0.21}  & \cellcolor[HTML]{EFEFEF} & \cellcolor[HTML]{EFEFEF}\textbf{6.76} & 1.69           & \cellcolor[HTML]{EFEFEF} & \cellcolor[HTML]{EFEFEF}\textbf{8.88}  & \cellcolor[HTML]{EFEFEF}\textbf{1.19} \\ \hline
\multirow{3}{*}{Yi1.5-9B}    & \textit{orig} &                 & 15.03         & 9.04          &  & 44.34          & 14.13         &  & 33.67          & 11.60         &  & 23.87          & 10.41           &  & 9.21          & 8.08           &  & 25.22          & 10.65         \\
                             & \textit{semt} &               & 6.17          & 1.94          &  & 12.86          & 2.31          &  & 25.50          & 3.79          &  & 6.48           & 1.89           &  & 2.53          & 1.71           &  & 10.71          & 2.33          \\
                             & \textit{ours} &                   & \cellcolor[HTML]{EFEFEF}\textbf{4.50} & \cellcolor[HTML]{EFEFEF}\textbf{0.51} & \cellcolor[HTML]{EFEFEF} & \cellcolor[HTML]{EFEFEF}\textbf{7.59}  & \cellcolor[HTML]{EFEFEF}\textbf{0.55} &\cellcolor[HTML]{EFEFEF}  & \cellcolor[HTML]{EFEFEF}\textbf{19.66} & \cellcolor[HTML]{EFEFEF}\textbf{2.00} & \cellcolor[HTML]{EFEFEF} & \cellcolor[HTML]{EFEFEF}\textbf{-3.50} & \cellcolor[HTML]{EFEFEF}\textbf{0.27} &\cellcolor[HTML]{EFEFEF}  & \cellcolor[HTML]{EFEFEF}\textbf{0.45}  & \cellcolor[HTML]{EFEFEF}\textbf{-0.37}  & \cellcolor[HTML]{EFEFEF} & \cellcolor[HTML]{EFEFEF}\textbf{5.74}  & \cellcolor[HTML]{EFEFEF}\textbf{0.59} \\ \hline
\multirow{3}{*}{Qwen2.5-7B} & \textit{orig} &                 & 6.65          & 3.44          &  & 19.77          & 4.86          &  & 18.51          & 5.24          &  & 8.06           & 4.16           &  & 7.08          & 6.03           &  & 12.01          & 4.74          \\
                             & \textit{semt} &               & 4.93          & 1.06          &  & 10.37          & 2.77          &  & 18.06          & 2.87          &  & 3.21           & 2.32           &  & \cellcolor[HTML]{EFEFEF}\textbf{1.74}           & \cellcolor[HTML]{EFEFEF}\textbf{0.72}  &  & 7.66           & 1.95          \\
                             & \textit{ours} &                   & \cellcolor[HTML]{EFEFEF}\textbf{4.72} & \cellcolor[HTML]{EFEFEF}\textbf{0.61} &\cellcolor[HTML]{EFEFEF}  & \cellcolor[HTML]{EFEFEF}\textbf{6.82}  & \cellcolor[HTML]{EFEFEF}\textbf{2.31} & \cellcolor[HTML]{EFEFEF} & \cellcolor[HTML]{EFEFEF}\textbf{15.25}  & \cellcolor[HTML]{EFEFEF}\textbf{2.32} & \cellcolor[HTML]{EFEFEF} & \cellcolor[HTML]{EFEFEF}\textbf{0.04}  & \cellcolor[HTML]{EFEFEF}\textbf{-0.39}  & \cellcolor[HTML]{EFEFEF} & 2.31  & 1.08           & \cellcolor[HTML]{EFEFEF} & \cellcolor[HTML]{EFEFEF}\textbf{5.83}  & \cellcolor[HTML]{EFEFEF}\textbf{1.19} \\ \hline
\multirow{3}{*}{Qwen2.5-14B}  & \textit{orig} &                 & 11.53         & 5.71          &  & 27.75          & 9.78          &  & 20.83          & 8.10          &  & 19.95          & 6.94           &  & 7.21          & 5.43           &  & 17.45          & 7.19          \\
                             & \textit{semt} &               & 5.79          & 1.46          &  & 1.46           & 2.76          &  & 17.03          & 2.87          &  & 5.49           & 1.42           &  & \cellcolor[HTML]{EFEFEF}\textbf{0.52}           & 0.12           &  & 6.06           & 1.72          \\
                             & \textit{ours} &                   & \cellcolor[HTML]{EFEFEF}\textbf{4.57} & \cellcolor[HTML]{EFEFEF}\textbf{0.99} &\cellcolor[HTML]{EFEFEF}  & \cellcolor[HTML]{EFEFEF}\textbf{-3.57} & \cellcolor[HTML]{EFEFEF}\textbf{0.93} & \cellcolor[HTML]{EFEFEF} & \cellcolor[HTML]{EFEFEF}\textbf{13.98}  & \cellcolor[HTML]{EFEFEF}\textbf{1.20} & \cellcolor[HTML]{EFEFEF} & \cellcolor[HTML]{EFEFEF}\textbf{-4.73} & \cellcolor[HTML]{EFEFEF}\textbf{0.37}  & \cellcolor[HTML]{EFEFEF} & 4.38  & \cellcolor[HTML]{EFEFEF}\textbf{0.00}  & \cellcolor[HTML]{EFEFEF} & \cellcolor[HTML]{EFEFEF}\textbf{2.93}  & \cellcolor[HTML]{EFEFEF}\textbf{0.70} \\ \hline
\end{tabular}
\caption{Data contamination resistance (\%) of eight open-source models across simulated scenarios. \textit{orig} denotes using original dataset, \textit{semt} denotes using semantic dataset, which involves restating the text while preserving its original meaning, and \textit{ours} denotes using our updated dataset. Following Section~\ref{sec-performance_test}, we use multiple prompt templates to mitigate prompt biases, reporting averaged performance. Best performances are in bold.}
\label{tab-mitigation_indicators}
\end{table*}

To assess the effectiveness of our method in mitigating the overestimation problem caused by data contamination, we follow prior studies~\cite{zhou2023don, ying2024automating} and simulate data contamination scenarios. 
Specifically, we introduce test prompts and the test set with ground truth labels, during the training phase to simulate data contamination conditions, enabling a rigorous assessment of our approach’s resistance to data leakage.

We conduct contamination simulations on eight open-source models, comparing results across three types of datasets: the original dataset $\mathcal{D}$, semantic dataset $\mathcal{D}^s$, and our updated dataset $ \mathcal{\hat D}$. Detailed training configurations are provided in Appendix~\ref{appendix-configuration}. The results are presented in Table~\ref{tab-mitigation_indicators}, where $\delta_1$ captures both the model's ability to improve task comprehension and its potential to memorize test set information due to contamination. 
In contrast, $\delta_2$  isolates the effect of training data, making it a more reliable indicator of contamination by attributing performance gains solely to test set exposure. This distinction ensures that $\delta_2$ provides a precise measure of an LLM’s resistance to data contamination.
Our observations reveal several critical trends regarding data contamination in LLMs:

\textbf{Performance overestimation intensifies with increasing model size in contaminated settings}. For instance, in our simulation using the original dataset, Qwen2.5-7B shows $\delta_1$ and $\delta_2$ values of 12.01 and 4.74, respectively, whereas the larger Qwen2.5-14B model exhibits higher values of 17.45 and 7.19. This trend is consistent across different model series. However, when tested on our updated dataset, these parameter-scale-induced discrepancies are significantly reduced.

\textbf{Cognitively complex tasks are more sensitive to data contamination}. Tasks such as irony detection, stance detection, and RTE, consistently yield higher $\delta$ values, suggesting a positive correlation between task cognitive complexity contamination sensitivity. These cognitively demanding tasks may prompt models to rely more on shortcuts like memorization, making them more vulnerable to data contamination compared to simpler tasks like emotion recognition and MPRC. 

\textbf{Our real-world knowledge integration method significantly improves contamination mitigation}.
While simple data rewriting techniques provide some resistance to data contamination, our method, incorporating real-world real-time knowledge, demonstrates superior performance mitigating overestimation and counteracting the effects of contamination. 
Notably, it outperforms conventional approaches such as \textit{semt}, highlighting the importance of dynamic knowledge updates in ensuring model robustness.

\begin{figure*}[!t]
  \centering
  \includegraphics[width=\linewidth]{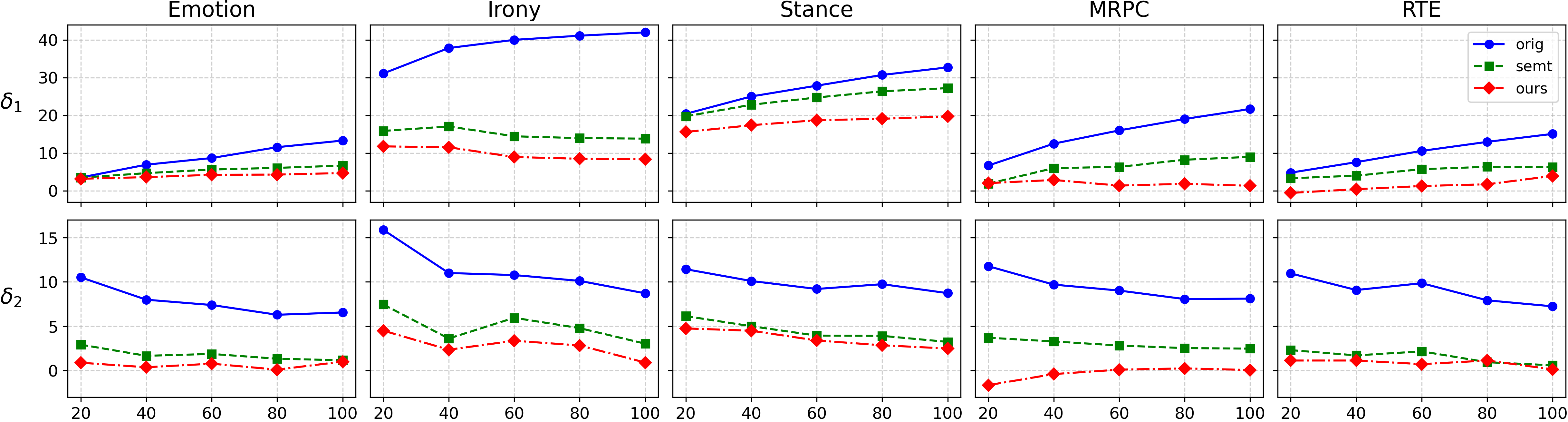}
  \caption{Data contamination resistance (\%) of eight open-source models under different data proportions (20\%, 40\%, 60\%, 80\%, 100\%). The first row shows $\delta_1$ values for the test set-only scenario across the original dataset, semantic dataset, and our updated dataset. The second row presents $\delta_2$ values for the train and test set scenario. The results are the mean values calculated across all eight open-source models.}
  \label{fig-ratio}
\end{figure*}

\begin{table*}[!t]
\small
\centering
\setlength{\tabcolsep}{4pt}
\renewcommand{\arraystretch}{1.1}
\begin{tabular}{lrrrrrrrrrrrrrrrrrrr}
\hline
\multirow{2}{*}{}       & \multirow{2}{*}{} & \multicolumn{2}{c}{Emotion}   &  & \multicolumn{2}{c}{Irony}      &  & \multicolumn{2}{c}{Stance}     &  & \multicolumn{2}{c}{MRPC}        &  & \multicolumn{2}{c}{RTE}         &  & \multicolumn{2}{c}{AVG}        \\ \cline{3-4} \cline{6-7} \cline{9-10} \cline{12-13} \cline{15-16} \cline{18-19} 
                             &                       & $\mathit{\delta_1} \downarrow$        & $\mathit{\delta_2} \downarrow$        &  & $\mathit{\delta_1} \downarrow$         & $\mathit{\delta_2} \downarrow$        &  & $\mathit{\delta_1} \downarrow$         & $\mathit{\delta_2} \downarrow$        &  & $\mathit{\delta_1} \downarrow$         & $\mathit{\delta_2} \downarrow$         &  & $\mathit{\delta_1} \downarrow$         & $\mathit{\delta_2} \downarrow$         &  & $\mathit{\delta_1} \downarrow$         & $\mathit{\delta_2} \downarrow$        \\ \hline
\multirow{1}{*}{Llama3-8B}  &                  & -0.32          & -0.01          &  & -0.13          & 0.12          &  & 0.03          & -0.06          &  & 0.22          & 0.05           &  & -9.20          & -4.77           &  & -1.88          & -0.93          \\
\multirow{1}{*}{Llama2-13B}   &                  & 0.23         & 0.05          &  & 0.28          & 0.83          &  & 0.00          & 0.05          &  & 0.24          & 0.15           &  & -4.57          & -3.72           &  & -0.76          & -0.53          \\
\multirow{1}{*}{Ministral-8B} &                  & -0.15         & -0.14          &  & -0.14          & 0.56          &  & 0.12          & 0.18          &  & 1.52          & -0.06           &  & -15.14          & -10.61           &  & -2.76          & -2.01          \\
\multirow{1}{*}{Mistral-12B}  &                  & 0.18         & 0.09          &  & 0.08          & -0.52         &  & -0.02          & 0.40          &  & 0.36          & 0.22           &  & -0.21          & 0.96           &  & 0.08          & 0.23          \\
\multirow{1}{*}{Yi1.5-6B}    &                  & -0.48         & 0.40          &  & 0.30          & 0.45          &  & 0.03          & 0.23          &  & -0.07          & -0.21           &  & -3.29          & 4.28           &  & -0.70          & 1.03          \\
\multirow{1}{*}{Yi1.5-9B}    &                  & 0.02         & 0.29          &  & -0.20          & 0.73         &  & -0.29          & -0.25         &  & -0.03          & 0.23           &  & -3.77          & -4.08           &  & -0.85          & -0.62         \\
\multirow{1}{*}{Qwen2.5-7B} &                  & 0.03          & 0.25          &  & 0.04          & 0.25          &  & 0.23          & -0.11          &  & 0.24           & 0.25           &  & -2.28          & -3.82           &  & -0.35          & -0.64          \\
\multirow{1}{*}{Qwen2.5-14B}  &                  & -0.06         & 0.10          &  & -0.05          & -0.09          &  & -0.17          & -0.19          &  & 0.30          & 0.15           &  & -1.40          & -2.71           &  & -0.28          & -0.55          \\
\hline

\end{tabular}
\caption{Data contamination resistance performance (\%) of eight open-source models on original datasets under text-only contamination scenarios.}
\label{tab-mitigation_indicators_incre}
\vspace{-5pt}
\end{table*}

\subsection{Impact of Contamination Proportion (Q3)}
In this section, we examine how varying data proportions influence the effects of data contamination.
For the \textit{`test set only'} simulated scenario, we sample different proportions of the test set to compute $\delta_1$ and analyze how varying ratios of the test data contamination impact performance overestimation. 
For the \textit{`training set and test set'} simulated scenario, we vary the proportion of the training set and compute $\delta_2$ by incorporating it with the test set. 
All training configurations remain consistent with those detailed in Section~\ref{sec-contamination_test}. The results are visualized in Figure~\ref{fig-ratio}.

\textbf{$\delta_1$ exhibits an upward trend, reflecting increasing performance overestimation as more test set data is exposed}. 
This is expected, as greater test set contamination amplifies the model’s memorization effect, artificially inflating performance.

\textbf{$\delta_2$ demonstrates a downward trend, aligning with the explanation} in Section~\ref{sec-contamination_test}. 
This metric isolates and quantifies performance improvements resulting from test set contamination, independent of enhanced task understanding. 
When incorporating the training set during the training process, models develop task understanding primarily through training data rather than test data. 
Therefore, as the proportion of the training set increases, $\delta_2$ effectively filters out the performance gains attributed to task understanding from test data, leading to a more precise measurement of performance overestimation due to contamination by the test data.

\textbf{Our updated dataset demonstrates stronger resistance to data contamination across both scenarios}, significantly reducing performance overestimation regardless of task complexity or the ratio between test and training sets. 
Further analysis of the mean and variance of $\delta_1$ and $\delta_2$ across different proportions for the original, semantic, and our datasets (outlined in Appendix~\ref{appendix-proportion}) reveals that our CoreEval provides more stable metrics across various data proportions compared to both the original and semantic datasets. These findings underscore the critical role of incorporating real-world and real-time knowledge into dataset design to enhance model robustness against data contamination.

\subsection{Impact of Contamination Types (Q3)}
In this section, we further extend our investigation by implementing a text-only contamination test, drawing upon the methodologies proposed by~\citet{li2024open} and~\citet{jiang2024investigating}. Diverging from previous simulation scenarios that involved the exposure of both test labels and texts during the training phase, this specific experimental setup exclusively leaks the textual content of the evaluation samples. 
Detailed training configurations are elaborated in Appendix~\ref{appendix-configuration}, and the comprehensive results are presented in Table~\ref{tab-mitigation_indicators_incre}.

The experimental findings indicate that the $\delta_1$ and $\delta_2$ values, when measured on the original datasets under these text-only contamination conditions, are predominantly negative across eight distinct open-source models. 
This observation suggests that text-only contamination, without label leakage, does not contribute to performance overestimation, consistent with the prior research by \citet{li2024open}. 
Conversely, the substantial performance improvements observed in Table~\ref{tab-mitigation_indicators}, where test sets including ground truth labels and test prompts are contaminated, highlight the critical need for targeted mitigation strategies to address this type of data contamination.

\section{Conclusion}
In this paper, we introduce CoreEval, an automatic contamination-resilient evaluation framework incorporating real-time real-world knowledge. We further propose a structured workflow engineered to guarantee the timeliness and reliability of LLM evaluations. Extensive experiments across various NLP tasks demonstrate CoreEval's robust effectiveness in mitigating data contamination. CoreEval is developed to be broadly applicable across NLP tasks, delivering efficient contamination-resilient evaluation while ensuring high data quality with minimal human intervention, thus facilitating fairer and more timely LLM assessment.

\section*{Acknowledgements}
This work was partially supported by the National Natural Science Foundation of China 62176076, Natural Science Foundation of Guang Dong 2023A1515012922, the Shenzhen Foundational Research Funding JCYJ20220818102415032, the Major Key Project of PCL2023A09, Guangdong Provincial Key Laboratory of Novel Security Intelligence Technologies 2022B1212010005 and CIPSC-SMP-ZHIPU Large Model Cross-Disciplinary Fund ZPCG20241119405.

\section*{Limitations}
Our proposed CoreEval framework updates text based on up-to-date and real-world knowledge. Although we have implemented data reflection and iteration processes to minimize inaccuracies, there is a possibility of generating a minimal amount of hallucinated data. Given our manual evaluation scores for the quality of updated data, the impact of such minimal hallucinated data on the evaluation of LLMs for most NLP tasks is negligible. Furthermore, in this study, CoreEval is applied only to classification tasks. In the future, we plan to extend its application to more complex tasks such as question answering and summarization.

\section*{Ethics Statement}
The datasets used in this study are sourced from open-access datasets, ensuring compliance with data accessibility standards. We have taken measures to remove any information related to user privacy from these datasets to protect individual identities and maintain confidentiality. The real-world knowledge required for updates is sourced from GDELT. While updating the data, there is a possibility of introducing references to relevant individuals or events. We have made every effort to ensure that these references are accurate and respectful.

\bibliography{custom}

\appendix

\section{Various Prompt Templates}
\label{appendix-prompt}
\subsection{Prompt of CoreEval Framework}
\label{appendix-prompt_framework}

\begin{figure*}[!h]
  \centering
  \includegraphics[width=0.5\linewidth]{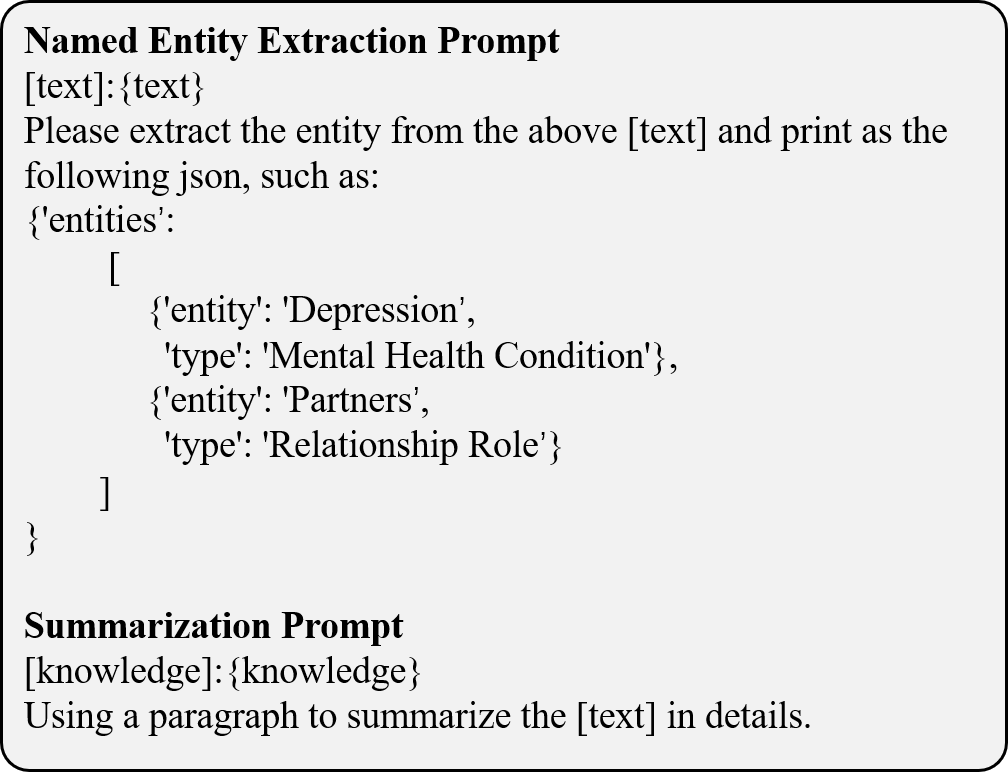}
  \caption{Prompt in Real-World Knowledge Attainment}
  \label{appendix-1}
\end{figure*}

\begin{figure*}[!h]
  \centering
  \includegraphics[width=0.5\linewidth]{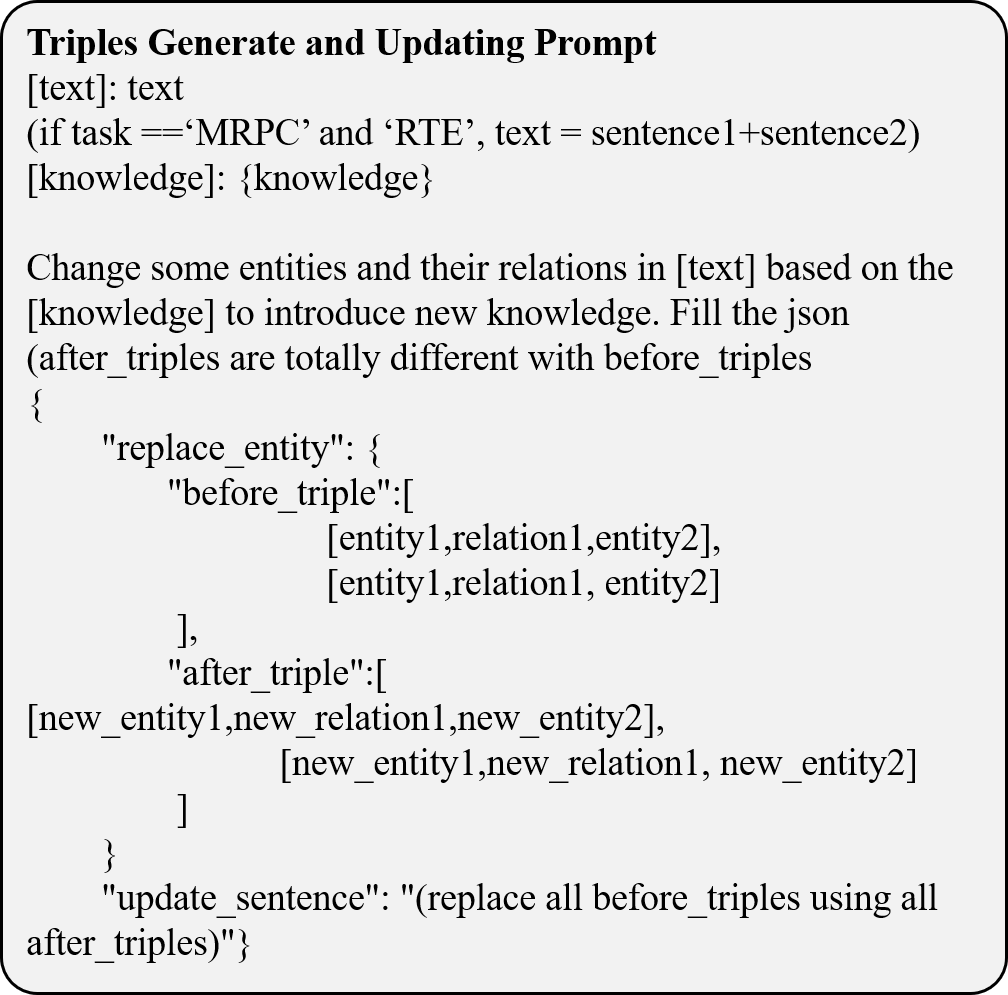}
  \caption{Prompt of triples generation and updating.}
  \label{appendix-2-1}
\end{figure*}

\begin{figure*}[t]
  \centering
  \includegraphics[width=\linewidth]{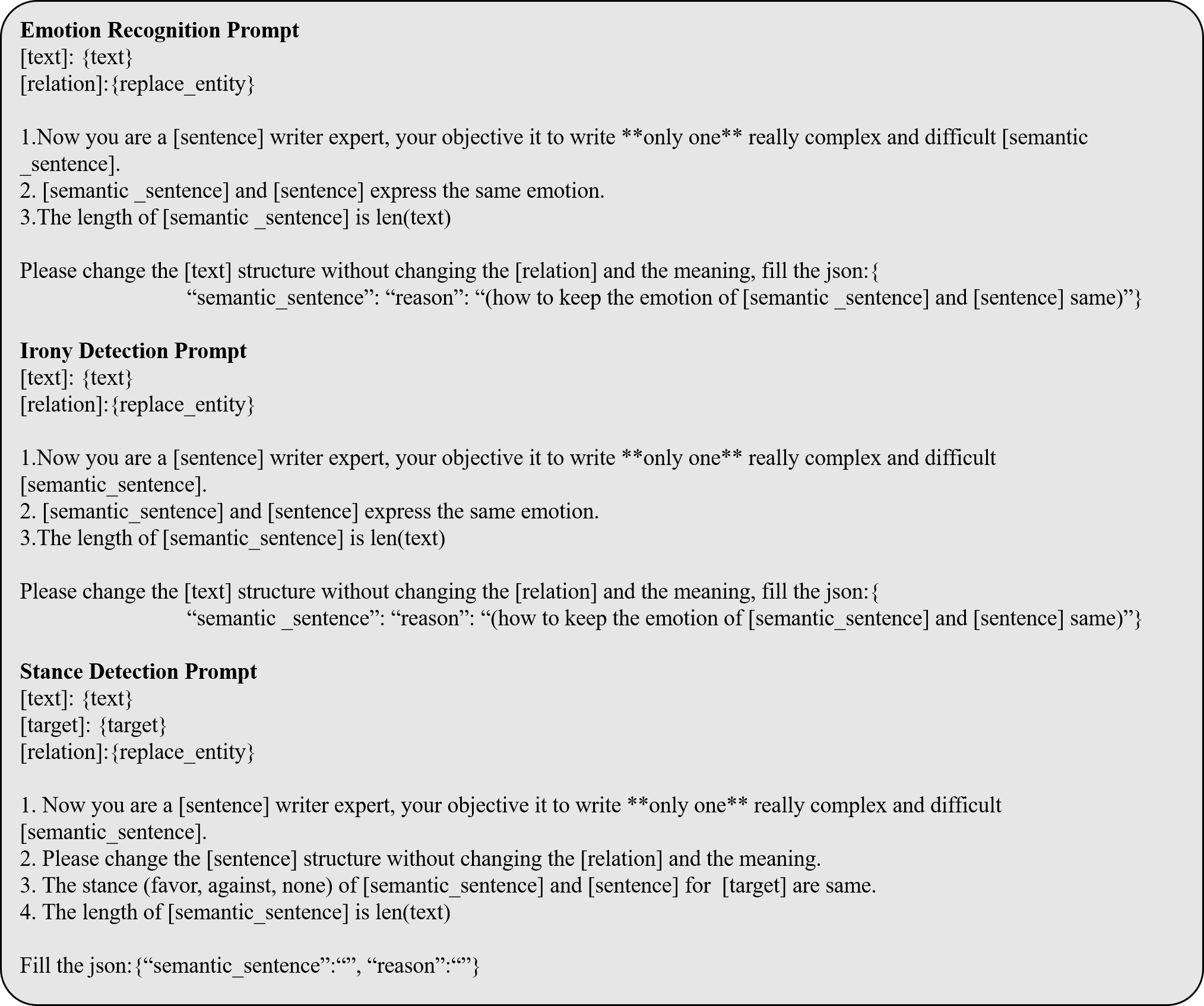}
  \caption{Prompt of semantic rewriting for emotion recognition, irony detection, and stance detection tasks.}
  \label{appendix-2-2}
\end{figure*}

\begin{figure*}[t]
  \centering
  \includegraphics[width=\linewidth]{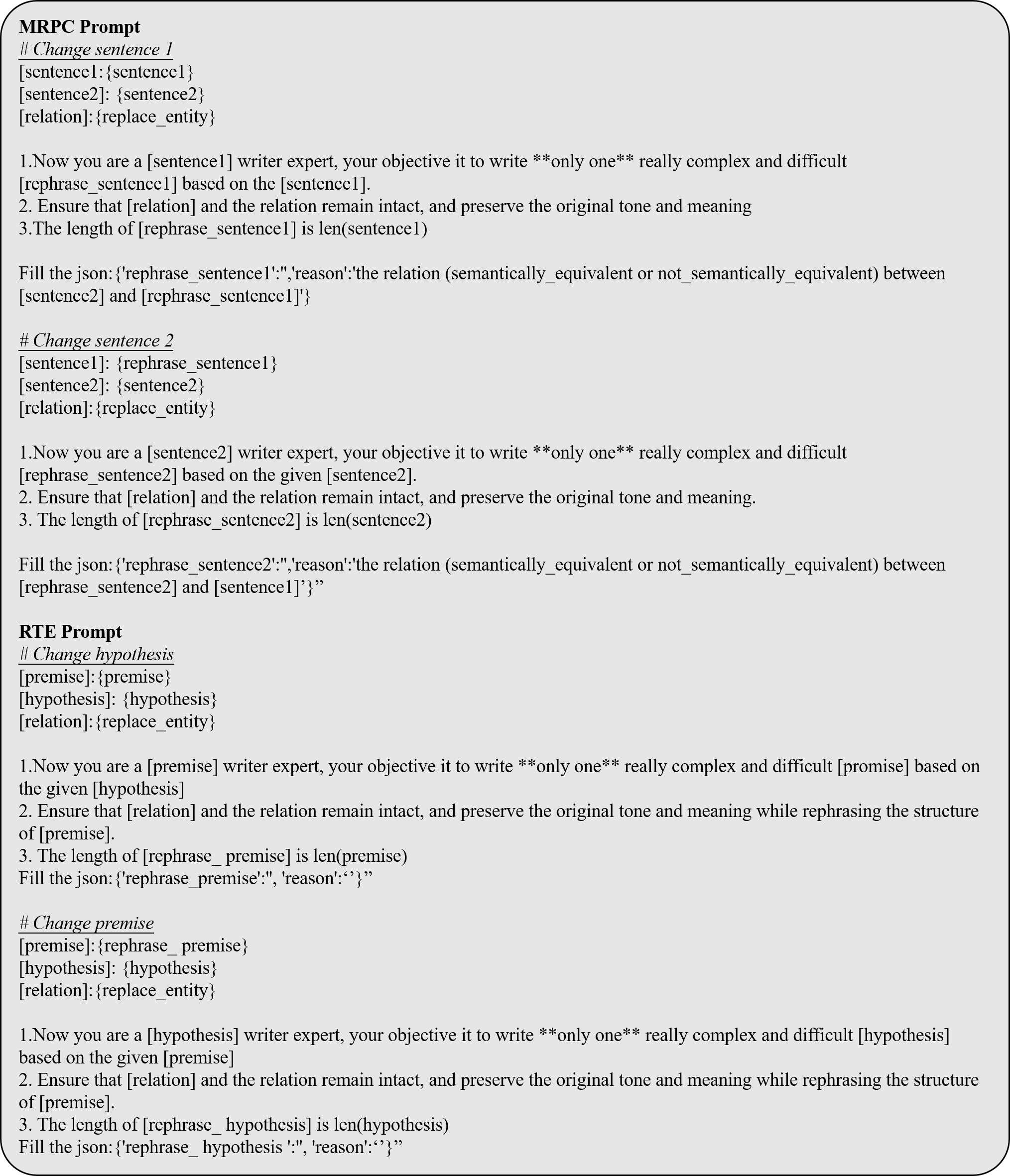}
  \caption{Prompt of semantic rewriting for MRPC and RTE tasks.}
  \label{appendix-2-3}
\end{figure*}

\begin{figure*}[t]
  \centering
  \includegraphics[width=\linewidth]{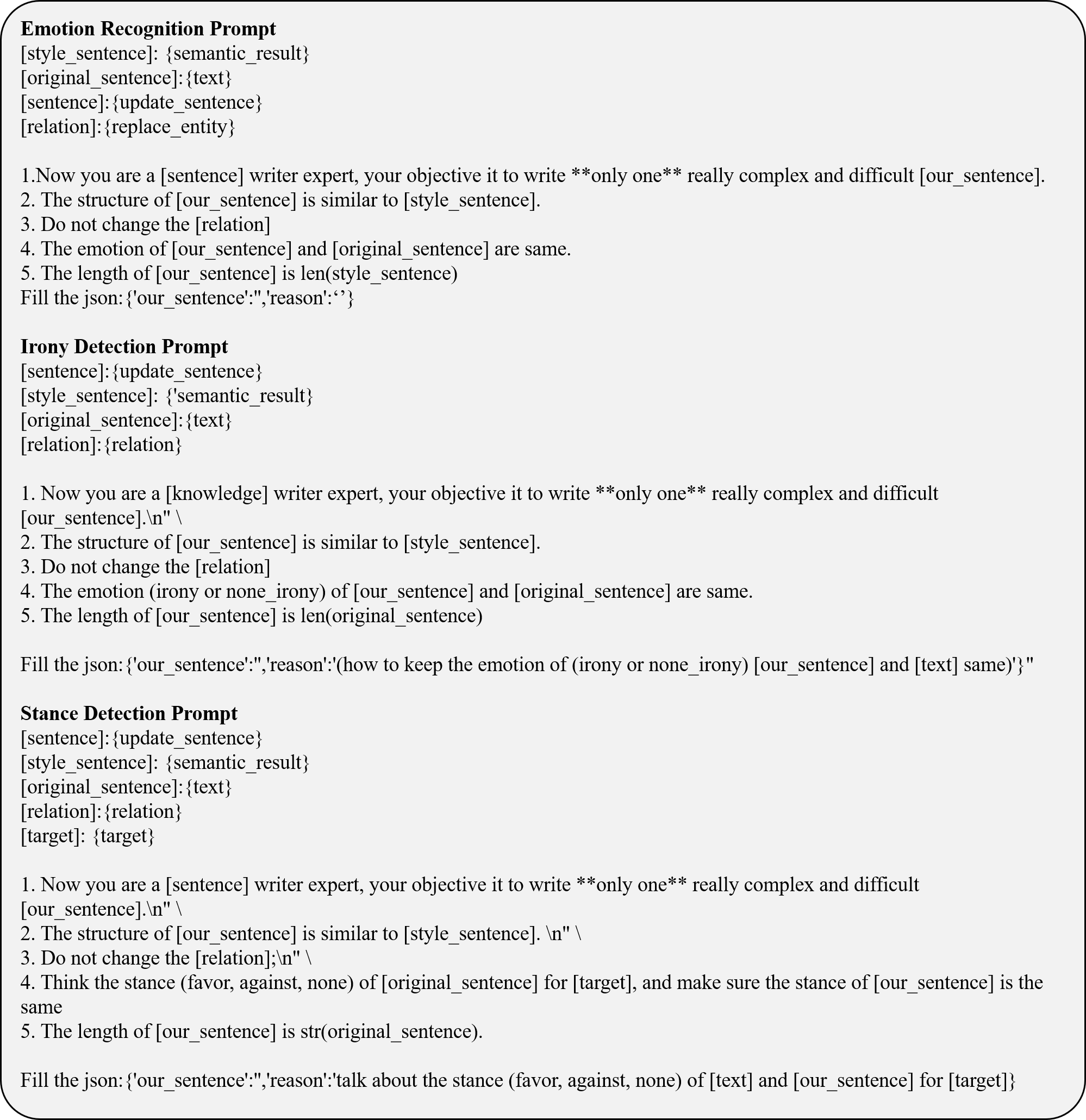}
  \caption{Prompt of updated sentence for emotion recognition, irony detection, and stance detection tasks.}
  \label{appendix-2-4}
\end{figure*}

\begin{figure*}[t]
  \centering
  \includegraphics[width=\linewidth]{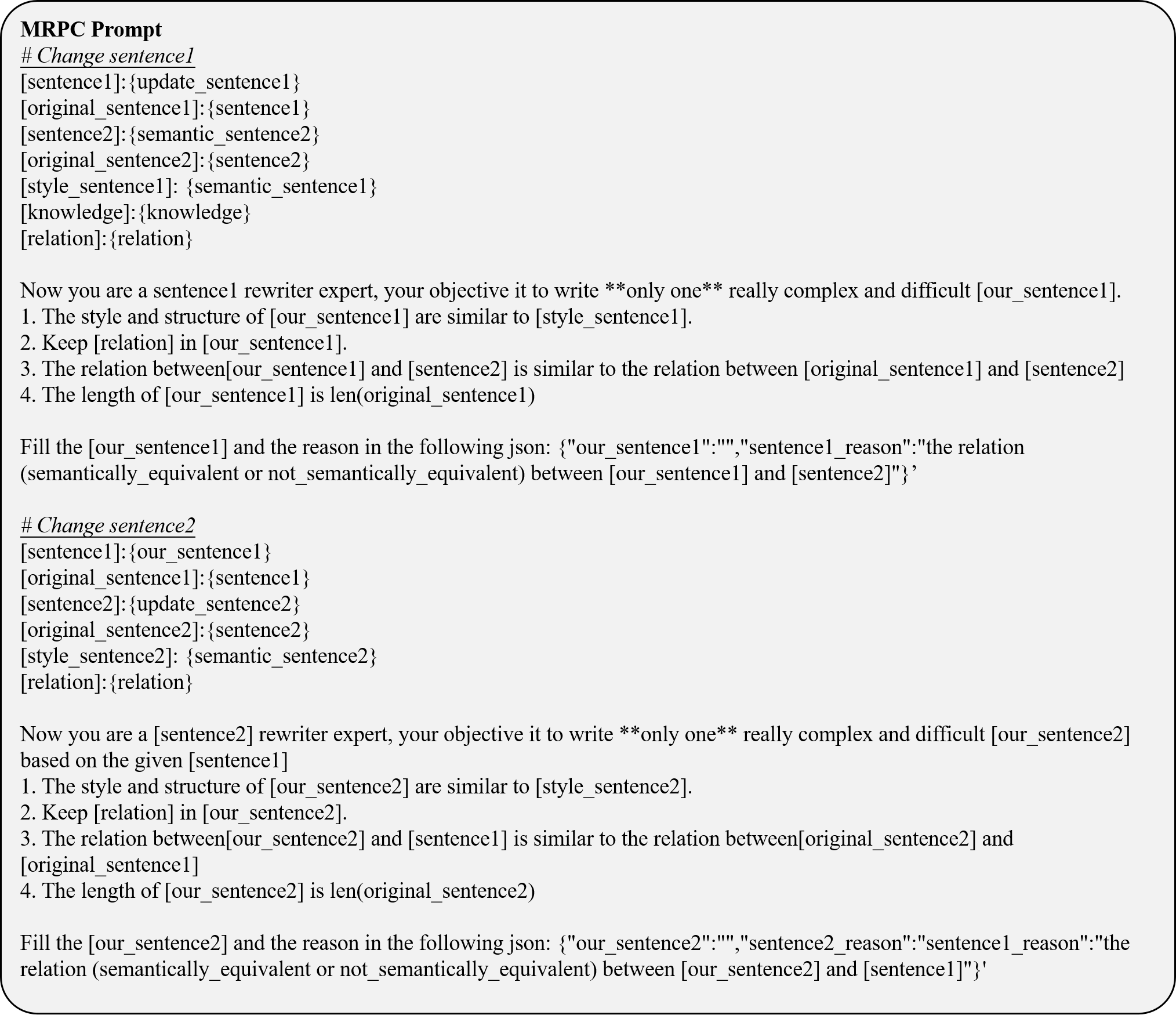}
  \caption{Prompt of semantic rewriting for MRPC task.}
  \label{appendix-2-5}
\end{figure*}

\begin{figure*}[t]
  \centering
  \includegraphics[width=\linewidth]{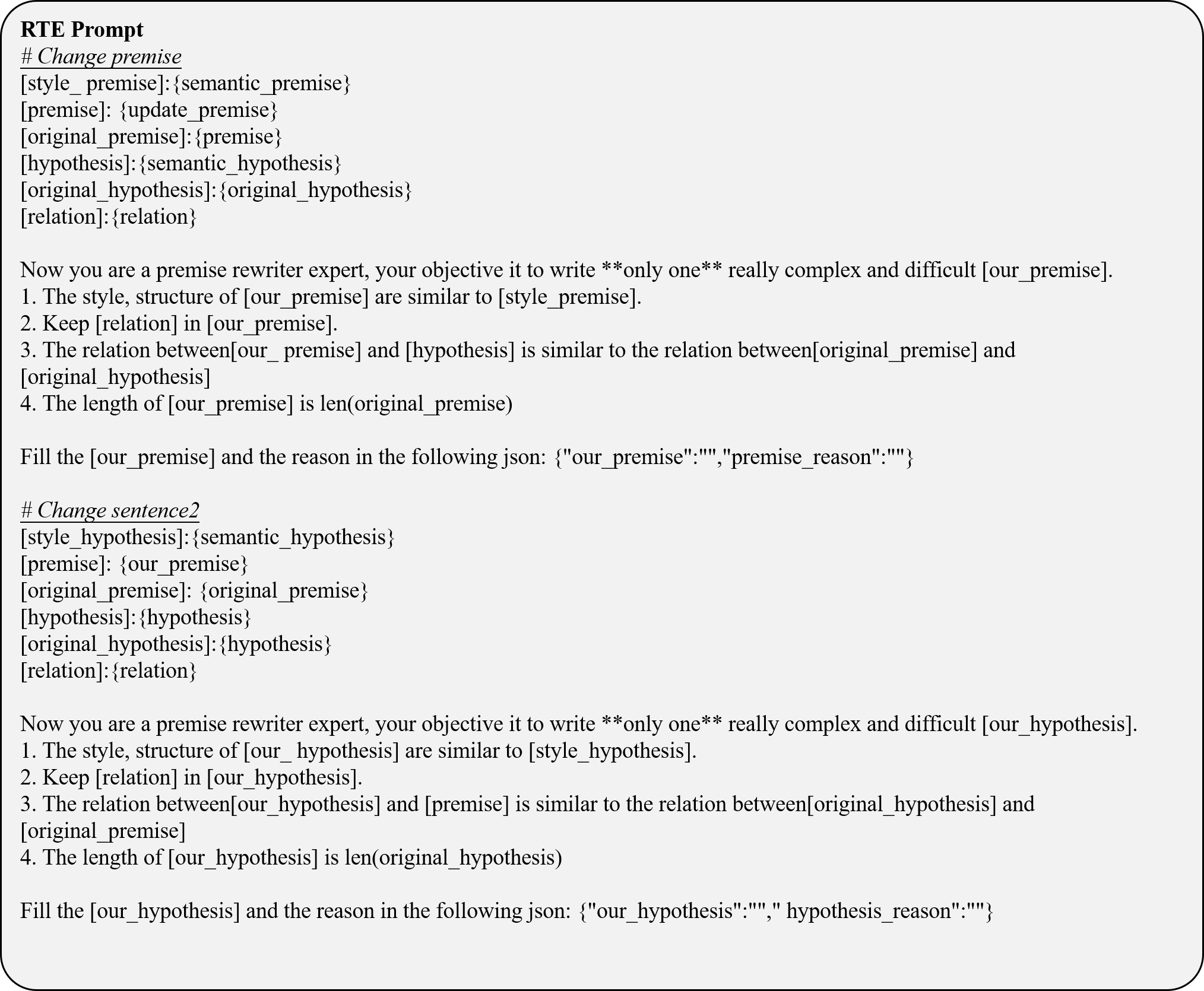}
  \caption{Prompt of semantic rewriting for RTE task.}
  \label{appendix-2-6}
\end{figure*}

\begin{figure*}[t]
  \centering
  \includegraphics[width=0.5\linewidth]{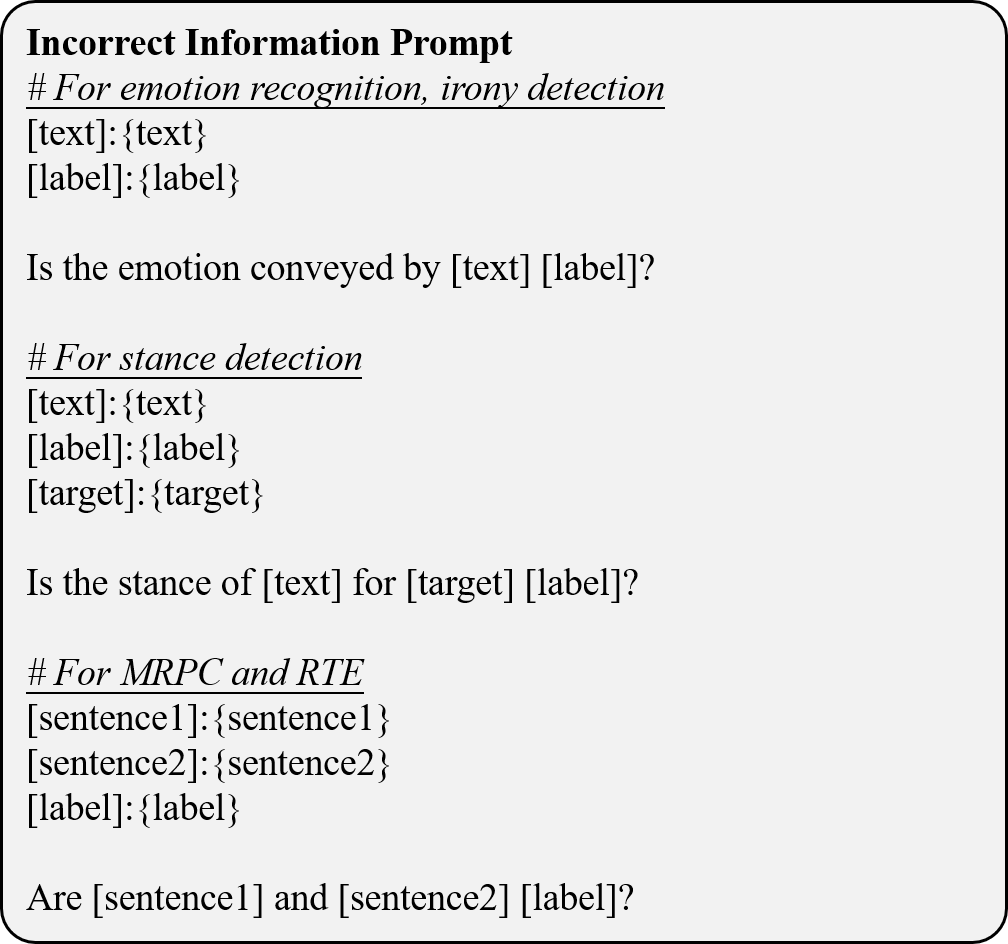}
  \caption{Prompt in Incorrect Information of Data Reflection}
  \label{appendix-3-1}
\end{figure*}

\begin{figure*}[t]
  \centering
  \includegraphics[width=0.5\linewidth]{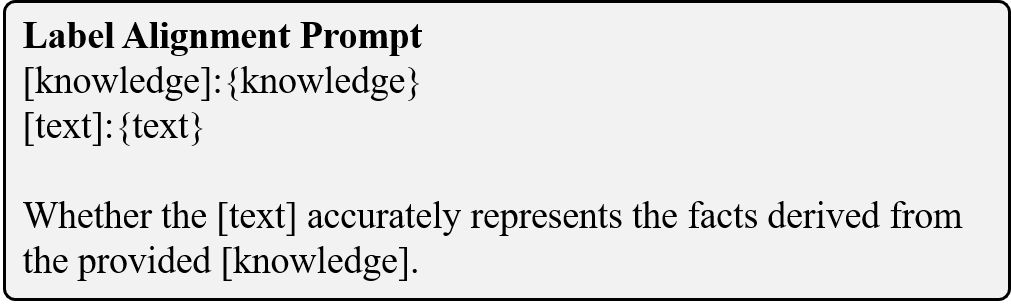}
  \caption{Prompt in Label Alignment of Data Reflection}
  \label{appendix-3-2}
\end{figure*}

Figure~\ref{appendix-1} presents the prompts in the process of Real-World Knowledge Attainment.
The workflows of Knowledge contextualization are shown in Figure~\ref{appendix-2-1}, Figure~\ref{appendix-2-2}, Figure~\ref{appendix-2-3}, Figure~\ref{appendix-2-4}, Figure~\ref{appendix-2-5}, and Figure~\ref{appendix-2-6}.
Ultimately, Figure~\ref{appendix-3-1} and Figure~\ref{appendix-3-2} demonstrate the prompts of data reflection.

\subsection{Prompt of Contamination Test}
\label{appendix-prompt_experiment}
To address potential result bias stemming from task sensitivity to prompts, we employed three prompt templates for each task. The performance metrics were then averaged across these prompt variations to obtain the final results. The comprehensive set of prompt templates utilized for all five tasks is detailed in Table~\ref{tab:emotion_prompts}, \ref{tab:irony_prompts}, \ref{tab:stance_prompts}, \ref{tab:mrpc_prompts}, and \ref{tab:rte_prompts}, which present the complete prompt formulations for each task-specific evaluation.

\section{Data Contamination Resistance Indicators}
\label{appendix-resistance_indicators}
Data contamination, which refers to the inflated performance of a model on a specific dataset or benchmark due to the leakage of test data, can distort the true evaluation and assessment of a LLM's capabilities.~\cite{zhou2023don, dekoninck2024constat} Therefore, mitigating the overestimation of performance caused by data contamination is key to addressing this issue. The degree of spurious performance growth following data contamination becomes the primary metric for evaluating data contamination mitigation efforts.

However, precisely determining whether a model has been contaminated by certain datasets remains challenging in practice. Previous studies have simulated data contamination by directly training models on test sets of specific datasets~\cite{ying2024automating, li2024open, jiang2024investigating, zhou2023don}. The mitigation effectiveness is then quantified by measuring the performance gap between the contaminated model before and after data updates. In our work, we similarly introduce $\delta_1$, which measures the performance difference between the model's evaluation results after training solely on the test set and its zero-shot performance (i.e., performance without any training) as one of the indicators for evaluating data contamination mitigation.

Furthermore, we argue that the performance improvements of LLMs directly exposed to test set data may stem from two sources: enhanced task understanding through exposure to task-specific data, and direct memorization effects from test set contamination. To isolate the latter effect, we propose $\delta_2$, which compares the performance difference between models trained on both train and test sets versus those trained exclusively on the train set. $\delta_2$ effectively eliminates the task-understanding gains from the train set while capturing the additional benefits derived from test set inclusion in training (i.e., the primary impact of data contamination), thereby providing a more accurate reflection of data contamination's contribution to model performance.

The substantial difference between these two indicators, as demonstrated in Table~\ref{tab-mitigation_indicators}, effectively validates this observation. Moreover, the declining trend of $\delta_2$ with increasing train set proportions, as illustrated in Figure~\ref{fig-ratio}, confirms that this indicator successfully isolates the impact of data contamination by removing the contribution of improved task understanding.

\section{Experimant Detail}
\subsection{Inference Configuration in Performance Test}
\label{appendix-configuration_inference}

For proprietary models, we set the temperature to 1.0, top-p to 1.0, max tokens to 1024, and fixed the seed to ensure experimental reproducibility.
For open-source models, we load model weights in bf16 format, set the temperature to 1.0, top-p to 1.0, max tokens to 512, and apply greedy decoding to guarantee reproducibility.

\subsection{Training Configuration in Contamination Test}
\label{appendix-configuration}

Due to computational resource constraints, we applied LoRA fine-tuning~\cite{hu2021lora} to eight open-source models. The LoRA hyperparameters were configured with a rank of 16, alpha of 32, dropout of 0.1, learning rate of 1e-4, and 3 epochs. For the RTE task, we set the training batch size to 2 and maximum sequence length to 512. For all other tasks, the maximum sequence length was set to 400, while the training batch size was adjusted according to model size. Specifically, Llama3-8B, Qwen2.5-7B, Mistral-8B, and Yi1.5-6B were trained with a batch size of 8; Yi1.5-9B, Llama2-13B, and Mistral-12B with a batch size of 3; and Qwen2.5-14B with a batch size of 2. For text-only contamination simulated scenarios, we configured the LoRA hyperparameters with a rank of 16, alpha of 32, dropout of 0.1, training batch size of 1, maximum sequence length of 1024, and 3 epochs. The learning rate was set to 1e-3 for the RTE task and 1e-5 for other tasks.

During inference, we employed a greedy decoding strategy by setting do\_sample to False and num\_sample to 1, thereby ensuring the reproducibility of our experimental results. 

\subsection{Experimental Result of Performance Test}
\label{appendix-performance}

We employed a greedy decoding strategy by setting do\_sample to False and num\_sample to 1, thereby ensuring the reproducibility of our experimental results. The detailed results of the original dataset and our updated dataset are presented in Table~\ref{tab-performance}.

\subsection{Experimental Result of Data Proportion Analysis}
\label{appendix-proportion}

Table~\ref{tab-proportion_result} presents the detailed experimental results of our data proportion analysis, encompassing the performance of eight open-source models across five tasks. The evaluation was conducted using varying proportions (20\%, 40\%, 60\%, 80\%, and 100\%) of both test and training sets, along with the average performance across all five tasks.

Table~\ref{tab-proportion_std} illustrates the standard deviations in data contamination resistance performance under varying data proportions for three datasets: the original, semantic, and our proposed updated dataset. The analysis reveals that our updated dataset consistently achieves lower variance compared to its counterparts. This reduced variability substantiates that our dataset yields more stable and robust evaluation metrics across different degrees of data contamination.

\section{Guideline of Human Evaluation}
\label{appendix-guideline}
Table \ref{tab-human} outlines the guidelines for human evaluation. Before presenting annotators with the final evaluation materials, we conduct a training session, providing them with this form and comprehensive instructions. This helps ensure they fully grasp the evaluation process, the significance of each metric, and the corresponding scoring standards.


\begin{table*}[!b]
\small
\centering
\begin{tabular}{p{0.12\textwidth}p{0.80\textwidth}}
\toprule
\textbf{Variant} & \textbf{Emotion Recognition Prompt} \\
\midrule
Prompt \#1 & Emotion detection is the task of identifying the emotional tone expressed in a given text. The possible emotions include joy, anger, sadness, and optimism.\newline
**The answer must be one and only one from the options: "joy", "anger", "sadness", and "optimism". No other responses are acceptable.**\newline
If none of the options seem to apply, select the one that is closest to the emotional tone of the text.\newline\newline
text: \{text\}\newline
Question: What is the primary emotion expressed in the text? Please select and only select the correct answer from "joy", "anger", "sadness", and "optimism". The response must strictly adhere to the following JSON FORMAT:\newline
\{\newline
    "emotion": "joy" | "anger" | "sadness" | "optimism"\newline
\} \\
\midrule
Prompt \#2 & Analyze the emotional tone of the following text. The emotions to choose from are joy, anger, sadness, and optimism.\newline
text: \{text\}\newline
Provide the detected emotion in JSON FORMAT:\newline
\{\newline
    "emotion": "joy" | "anger" | "sadness" | "optimism"\newline
\} \\
\midrule
Prompt \#3 & Emotion detection is the task of identifying the emotional tone expressed in a given text. The possible emotions include joy, anger, sadness, and optimism.\newline
**The answer must be one and only one from the options: "joy", "anger", "sadness", and "optimism". No other responses are acceptable.**\newline
If none of the options seem to apply, select the one that is closest to the emotional tone of the text.\newline\newline
text: \{text\}\newline
Question: What is the primary emotion expressed in the text? Please select and only select the correct answer from "joy", "anger", "sadness", and "optimism". \\
\bottomrule
\end{tabular}
\caption{Prompt templates for Emotion Recognition task.}
\label{tab:emotion_prompts}
\end{table*}

\begin{table*}[t]
\small
\centering
\begin{tabular}{p{0.12\textwidth}p{0.80\textwidth}}
\toprule
\textbf{Variant} & \textbf{Irony Detection Prompt} \\
\midrule
Prompt \#1 & Irony detection is the task of identifying whether a given text contains irony. Irony is when the literal meaning of the text is opposite or significantly different from the intended meaning, often used to convey criticism, sarcasm, or humor. The possible labels are "irony" and "not irony".\newline
text: \{text\}\newline
Question: Does the text contain irony? Irony can be characterized by:\newline
- A contrast between what is said and what is meant (verbal irony).\newline
- A situation where the expected outcome is different from the actual result (situational irony).\newline
- Exaggeration or sarcasm used to convey a hidden message or criticism.\newline
Please select the correct answer from "irony" and "not irony".\newline
Answer this question with JSON FORMAT:\newline
\{\newline
    "irony detection": "irony" | "not irony"\newline
\} \\
\midrule
Prompt \#2 & Analyze the following text to determine whether it contains irony. Irony is when there is a discrepancy between what is said and what is meant, or when the outcome of a situation is unexpected or opposite to what was anticipated. The labels to choose from are "irony" and "not irony".\newline
text: \{text\}\newline
Please provide your answer in JSON FORMAT:\newline
\{\newline
    "irony detection": "irony" | "not irony"\newline
\} \\
\midrule
Prompt \#3 & Irony detection is the task of identifying whether a given text contains irony. Irony often involves:\newline
- Saying one thing while meaning another (verbal irony).\newline
- A situation where the outcome is surprising or opposite to what was expected (situational irony).\newline
- Sarcasm or exaggerated statements to make a point or criticism.\newline
text: \{text\}\newline
Question: Does the text contain irony? Please select the correct answer from "irony" and "not irony". \\
\bottomrule
\end{tabular}
\caption{Prompt templates for Irony Detection task.}
\label{tab:irony_prompts}
\end{table*}

\begin{table*}[t]
\small
\centering
\begin{tabular}{p{0.12\textwidth}p{0.80\textwidth}}
\toprule
\textbf{Variant} & \textbf{Stance Detection Prompt} \\
\midrule
Prompt \#1 & Stance detection is to determine the attitude or tendency towards a certain target through a given sentence, including favor, against and neutral.\newline
text: \{text\}\newline
Question: What is the attitude of the text toward "\{target\}"? Please select the correct answer from "favor", "against" and "neutral".\newline
Answer this question with JSON FORMAT:\newline
\{\newline
    "stance": "favor" | "against" | "neutral"\newline
\} \\
\midrule
Prompt \#2 & Given a text, determine the sentiment towards the specified target: \{target\}. Possible answers are "favor", "against", or "neutral".\newline
text: \{text\}\newline
Please provide your answer in JSON FORMAT:\newline
\{\newline
    "stance": "favor" | "against" | "neutral"\newline
\} \\
\midrule
Prompt \#3 & Stance detection is to determine the attitude or tendency towards a certain target through a given sentence, including favor, against and neutral.\newline
text: \{text\}\newline
Question: What is the attitude of the text toward "\{target\}"? Please select the correct answer from "favor", "against" and "neutral". \\
\bottomrule
\end{tabular}
\caption{Prompt templates for Stance Detection task.}
\label{tab:stance_prompts}
\end{table*}

\begin{table*}[t]
\small
\centering
\begin{tabular}{p{0.12\textwidth}p{0.80\textwidth}}
\toprule
\textbf{Variant} & \textbf{MRPC Prompt} \\
\midrule
Prompt \#1 & The task is to determine whether a given sentence pair is semantically equivalent. The possible labels are "semantically equivalent" and "not semantically equivalent".\newline
sentence1: \{sentence1\}\newline
sentence2: \{sentence2\}\newline
Question: Are the sentences semantically equivalent? Please select the correct answer from "semantically equivalent" and "not semantically equivalent".\newline
Answer this question with JSON FORMAT:\newline
\{\newline
    "mrpc": "semantically equivalent" | "not semantically equivalent"\newline
\} \\
\midrule
Prompt \#2 & Given a pair of sentences, determine whether they are semantically equivalent. The possible labels are "semantically equivalent" and "not semantically equivalent".\newline
sentence1: \{sentence1\}\newline
sentence2: \{sentence2\}\newline
Please provide your answer in JSON FORMAT:\newline
\{\newline
    "mrpc": "semantically equivalent" | "not semantically equivalent"\newline
\} \\
\midrule
Prompt \#3 & The task is to determine whether a given sentence pair is semantically equivalent. The possible labels are "semantically equivalent" and "not semantically equivalent".\newline
sentence1: \{sentence1\}\newline
sentence2: \{sentence2\}\newline
Question: Are the sentences semantically equivalent? Please select the correct answer from "semantically equivalent" and "not semantically equivalent". \\
\bottomrule
\end{tabular}
\caption{Prompt templates for MRPC (Microsoft Research Paraphrase Corpus) task.}
\label{tab:mrpc_prompts}
\end{table*}

\begin{table*}[t]
\small
\centering
\begin{tabular}{p{0.12\textwidth}p{0.80\textwidth}}
\toprule
\textbf{Variant} & \textbf{RTE Prompt} \\
\midrule
Prompt \#1 & Recognizing Textual Entailment (RTE) is the task of determining whether a given premise entails a hypothesis. The possible labels are "entailment" and "not entailment".\newline
premise: \{premise\}\newline
hypothesis: \{hypothesis\}\newline
Question: Does the premise entail the hypothesis? Please select the correct answer from "entailment" and "not entailment".\newline
Answer this question with JSON FORMAT:\newline
\{\newline
    "rte": "entailment" | "not entailment"\newline
\} \\
\midrule
Prompt \#2 & Given a premise and a hypothesis, determine whether the premise entails the hypothesis. The possible labels are "entailment" and "not entailment".\newline
premise: \{premise\}\newline
hypothesis: \{hypothesis\}\newline
Please provide your answer in JSON FORMAT:\newline
\{\newline
    "rte": "entailment" | "not entailment"\newline
\} \\
\midrule
Prompt \#3 & Recognizing Textual Entailment (RTE) is the task of determining whether a given premise entails a hypothesis. The possible labels are "entailment" and "not entailment".\newline
premise: \{premise\}\newline
hypothesis: \{hypothesis\}\newline
Question: Does the premise entail the hypothesis? Please select the correct answer from "entailment" and "not entailment". \\
\bottomrule
\end{tabular}
\caption{Prompt templates for RTE (Recognizing Textual Entailment) task.}
\label{tab:rte_prompts}
\end{table*}

\begin{table*}[t]
\small
\centering
\setlength{\tabcolsep}{3.5pt}
\renewcommand{\arraystretch}{1.1}
\begin{tabular}{llcccccc}
\toprule
Model & & Emotion & Irony & Stance & MRPC & RTE & AVG \\ 
\midrule
\multirow{2}{*}{Llama3-8B} & \textit{orig} & 74.91 & 64.02 & 61.34 & 72.47 & 75.98 & 69.74 \\
 & \textit{ours} & 62.42 & 65.27 & 46.19 & 79.62 & 82.19 & 67.14 \\ \midrule
\multirow{2}{*}{Llama2-13B} & \textit{orig} & 74.25 & 43.29 & 47.68 & 57.86 & 65.02 & 57.62 \\
 & \textit{ours} & 58.06 & 47.24 & 36.41 & 72.06 & 67.20 & 56.19 \\ \midrule
\multirow{2}{*}{Mistral-8B} & \textit{orig} & 76.94 & 56.72 & 56.52 & 64.90 & 82.92 & 67.60 \\
 & \textit{ours} & 60.95 & 62.92 & 39.38 & 75.29 & 85.84 & 64.88 \\ \midrule
\multirow{2}{*}{Mistral-12B} & \textit{orig} & 73.96 & 56.95 & 55.95 & 67.82 & 83.91 & 67.72 \\
 & \textit{ours} & 57.55 & 63.53 & 39.09 & 78.61 & 85.30 & 64.82 \\ \midrule
\multirow{2}{*}{Qwen2.5-7B} & \textit{orig} & 74.43 & 71.15 & 61.02 & 70.30 & 87.59 & 72.90 \\
 & \textit{ours} & 60.01 & 67.13 & 44.93 & 84.13 & 83.48 & 67.94 \\ \midrule
\multirow{2}{*}{Qwen2.5-14B} & \textit{orig} & 78.19 & 69.12 & 68.08 & 72.95 & 88.93 & 75.45 \\
 & \textit{ours} & 60.85 & 70.05 & 46.55 & 83.40 & 83.69 & 68.91 \\ \midrule
\multirow{2}{*}{Yi1.5-6B} & \textit{orig} & 75.25 & 54.57 & 51.03 & 70.21 & 79.15 & 66.04 \\
 & \textit{ours} & 61.30 & 50.14 & 40.66 & 81.81 & 81.43 & 63.07 \\ \midrule
\multirow{2}{*}{Yi1.5-9B} & \textit{orig} & 77.48 & 53.40 & 59.36 & 72.48 & 87.89 & 70.12 \\
 & \textit{ours} & 61.73 & 56.45 & 40.91 & 81.75 & 85.54 & 65.28 \\ \midrule
\multirow{2}{*}{ChatGPT} & \textit{orig} & 78.32 & 62.93 & 65.67 & 70.86 & 83.53 & 72.26 \\
 & \textit{ours} & 63.23 & 62.54 & 49.29 & 83.36 & 82.88 & 68.26 \\ \midrule
\multirow{2}{*}{Gemini1.5} & \textit{orig} & 78.49 & 66.82 & 66.58 & 71.80 & 89.99 & 74.74 \\
 & \textit{ours} & 61.50 & 68.03 & 46.98 & 81.83 & 85.67 & 68.80 \\ \midrule
\multirow{2}{*}{Claude3.5} & \textit{orig} & 77.59 & 67.68 & 73.97 & 67.16 & 82.00 & 73.68 \\
 & \textit{ours} & 62.36 & 67.45 & 48.76 & 80.62 & 77.60 & 67.36 \\ 
\bottomrule
\end{tabular}
\caption{Performance (\%) of the eleven involved LLMs (zero-shot) on the original and our updated datasets. We utilize macro F1-score as the unified evaluation metric.}
\label{tab-performance}
\end{table*}

\begin{table*}[t]
\small
\centering
\setlength{\tabcolsep}{3.5pt}
\renewcommand{\arraystretch}{1.1}
\begin{tabular}{clccccccccccccccccccc}
\hline
\multirow{2}{*}{Proportion(\%)}       & \multirow{2}{*}{} & & \multicolumn{2}{c}{Emotion}   &  & \multicolumn{2}{c}{Irony}      &  & \multicolumn{2}{c}{Stance}     &  & \multicolumn{2}{c}{MRPC}        &  & \multicolumn{2}{c}{RTE}         &  & \multicolumn{2}{c}{AVG}        \\ \cline{4-5} \cline{7-8} \cline{10-11} \cline{13-14} \cline{16-17} \cline{19-20} 
                             &  &                       & $\mathit{\delta_1} \downarrow$        & $\mathit{\delta_2} \downarrow$        &  & $\mathit{\delta_1} \downarrow$         & $\mathit{\delta_2} \downarrow$        &  & $\mathit{\delta_1} \downarrow$         & $\mathit{\delta_2} \downarrow$        &  & $\mathit{\delta_1} \downarrow$         & $\mathit{\delta_2} \downarrow$         &  & $\mathit{\delta_1} \downarrow$         & $\mathit{\delta_2} \downarrow$         &  & $\mathit{\delta_1} \downarrow$         & $\mathit{\delta_2} \downarrow$        \\ \hline
\multirow{3}{*}{20}  & \textit{orig} &                 & 3.52          & 10.51          &  & 31.09          & 15.87          &  & 20.39          & 11.43          &  & 6.69          & 11.77           &  & 4.80          & 10.95           &  & 13.30          & 12.10          \\
                             & \textit{semt} &               & 3.48          & 2.92          &  & 15.86           & 7.45          &  & 19.73          & 6.14          &  & \cellcolor[HTML]{EFEFEF}\textbf{1.87}           & 3.70           &  & 3.32          & 2.30           &  & 8.85          & 4.50          \\
                             & \textit{ours} &                   & \cellcolor[HTML]{EFEFEF}\textbf{3.18} & \cellcolor[HTML]{EFEFEF}\textbf{0.88} & \cellcolor[HTML]{EFEFEF} & \cellcolor[HTML]{EFEFEF}\textbf{11.77}  & \cellcolor[HTML]{EFEFEF}\textbf{4.50} & \cellcolor[HTML]{EFEFEF} & \cellcolor[HTML]{EFEFEF}\textbf{15.58}  & \cellcolor[HTML]{EFEFEF}\textbf{4.75} & \cellcolor[HTML]{EFEFEF} & 2.03  & \cellcolor[HTML]{EFEFEF}\textbf{-1.65}  & \cellcolor[HTML]{EFEFEF} & \cellcolor[HTML]{EFEFEF}\textbf{-0.57}  & \cellcolor[HTML]{EFEFEF}\textbf{1.13}  & \cellcolor[HTML]{EFEFEF} & \cellcolor[HTML]{EFEFEF}\textbf{6.40}  & \cellcolor[HTML]{EFEFEF}\textbf{1.92} \\ \hline
\multirow{3}{*}{40}   & \textit{orig} &                 & 6.91         & 7.99          &  & 37.84          & 11.00          &  & 25.00          & 10.10          &  & 12.49          & 9.69           &  & 7.58          & 9.08           &  & 17.96          & 9.57          \\
                             & \textit{semt} &               & 4.67          & 1.66          &  & 17.05          & 3.61          &  & 22.80          & 5.00          &  & 6.00          & 3.28           &  & 4.00          & 1.71           &  & 10.90          & 3.05          \\
                             & \textit{ours} &                   & \cellcolor[HTML]{EFEFEF}\textbf{3.61} & \cellcolor[HTML]{EFEFEF}\textbf{0.37} & \cellcolor[HTML]{EFEFEF} & \cellcolor[HTML]{EFEFEF}\textbf{11.53} & \cellcolor[HTML]{EFEFEF}\textbf{2.35} & \cellcolor[HTML]{EFEFEF} & \cellcolor[HTML]{EFEFEF}\textbf{17.38} & \cellcolor[HTML]{EFEFEF}\textbf{4.49} & \cellcolor[HTML]{EFEFEF} & \cellcolor[HTML]{EFEFEF}\textbf{2.86} & \cellcolor[HTML]{EFEFEF}\textbf{-0.40}  & \cellcolor[HTML]{EFEFEF} & \cellcolor[HTML]{EFEFEF}\textbf{0.41} & \cellcolor[HTML]{EFEFEF}\textbf{1.13}  & \cellcolor[HTML]{EFEFEF} & \cellcolor[HTML]{EFEFEF}\textbf{7.16} & \cellcolor[HTML]{EFEFEF}\textbf{1.59} \\ \hline
\multirow{3}{*}{60} & \textit{orig} &                 & 8.68         & 7.39          &  & 40.00          & 10.77          &  & 27.86          & 9.20          &  & 16.00          & 9.02           &  & 10.57          & 9.85           &  & 20.62          & 9.25          \\
                             & \textit{semt} &               & 5.63          & 1.87          &  & 14.44          & 5.95          &  & 24.73          & 3.94          &  & 6.34           & 2.82           &  & 5.72          & 2.16           &  & 11.37          & 3.35          \\
                             & \textit{ours} &                   & \cellcolor[HTML]{EFEFEF}\textbf{4.23} & \cellcolor[HTML]{EFEFEF}\textbf{0.77} & \cellcolor[HTML]{EFEFEF} & \cellcolor[HTML]{EFEFEF}\textbf{8.95}  & \cellcolor[HTML]{EFEFEF}\textbf{3.36} & \cellcolor[HTML]{EFEFEF} & \cellcolor[HTML]{EFEFEF}\textbf{18.70} & \cellcolor[HTML]{EFEFEF}\textbf{3.39} & \cellcolor[HTML]{EFEFEF} & \cellcolor[HTML]{EFEFEF}\textbf{1.37}  & \cellcolor[HTML]{EFEFEF}\textbf{0.10}  & \cellcolor[HTML]{EFEFEF} & \cellcolor[HTML]{EFEFEF}\textbf{1.26} & \cellcolor[HTML]{EFEFEF}\textbf{0.72} & \cellcolor[HTML]{EFEFEF} & \cellcolor[HTML]{EFEFEF}\textbf{6.90}  & \cellcolor[HTML]{EFEFEF}\textbf{1.67} \\ \hline
\multirow{3}{*}{80}  & \textit{orig} &                 & 11.55         & 6.29          &  & 41.11          & 10.11         &  & 30.70          & 9.74          &  & 19.03          & 8.06           &  & 12.96          & 7.92           &  & 23.07          & 8.42          \\
                             & \textit{semt} &               & 6.07         & 1.33 &  & 13.96           & 4.78          &  & 26.36          & 3.91          &  & 8.22           & 2.53           &  & 6.35          & \cellcolor[HTML]{EFEFEF}\textbf{0.94}  &  & 12.19          & 2.70          \\
                             & \textit{ours} &                   & \cellcolor[HTML]{EFEFEF}\textbf{4.29} & \cellcolor[HTML]{EFEFEF}\textbf{0.10} & \cellcolor[HTML]{EFEFEF} & \cellcolor[HTML]{EFEFEF}\textbf{8.48}  & \cellcolor[HTML]{EFEFEF}\textbf{2.83} & \cellcolor[HTML]{EFEFEF} & \cellcolor[HTML]{EFEFEF}\textbf{19.08} & \cellcolor[HTML]{EFEFEF}\textbf{2.85} & \cellcolor[HTML]{EFEFEF} & \cellcolor[HTML]{EFEFEF}\textbf{1.85}  & \cellcolor[HTML]{EFEFEF}\textbf{0.24}  & \cellcolor[HTML]{EFEFEF} & \cellcolor[HTML]{EFEFEF}\textbf{1.72}  & 1.13           & \cellcolor[HTML]{EFEFEF} & \cellcolor[HTML]{EFEFEF}\textbf{7.09}  & \cellcolor[HTML]{EFEFEF}\textbf{1.43} \\ \hline
\multirow{3}{*}{100}    & \textit{orig} &                 & 13.31         & 6.55          &  & 41.98          & 8.71          &  & 32.71          & 8.73          &  & 21.68          & 8.11           &  & 15.08          & 7.24           &  & 24.95          & 7.87          \\
                             & \textit{semt} &               & 6.67          & 1.16          &  & 13.82          & 3.03          &  & 27.21          & 3.23          &  & 8.99           & 2.47           &  & 6.26          & 0.59  &  & 12.59          & 2.10          \\
                             & \textit{ours} &                   & \cellcolor[HTML]{EFEFEF}\textbf{4.70} & \cellcolor[HTML]{EFEFEF}\textbf{1.01} & \cellcolor[HTML]{EFEFEF} & \cellcolor[HTML]{EFEFEF}\textbf{8.34} & \cellcolor[HTML]{EFEFEF}\textbf{0.90} & \cellcolor[HTML]{EFEFEF} & \cellcolor[HTML]{EFEFEF}\textbf{19.72} & \cellcolor[HTML]{EFEFEF}\textbf{2.48} & \cellcolor[HTML]{EFEFEF} & \cellcolor[HTML]{EFEFEF}\textbf{1.32}  & \cellcolor[HTML]{EFEFEF}\textbf{0.06}  & \cellcolor[HTML]{EFEFEF} & \cellcolor[HTML]{EFEFEF}\textbf{3.93} & \cellcolor[HTML]{EFEFEF}\textbf{0.13}  & \cellcolor[HTML]{EFEFEF} & \cellcolor[HTML]{EFEFEF}\textbf{7.60}  & \cellcolor[HTML]{EFEFEF}\textbf{0.92} \\ \hline
\end{tabular}
\caption{Data contamination resistance performance (\%) of eight open-source models across simulated scenarios under different data
proportions (20\%, 40\%, 60\%, 80\%, 100\%). The results are the mean values calculated across all eight open-source models. \textit{orig} denote using original dataset, \textit{semt} denote using semantic dataset, and \textit{ours} denote using our updated dataset. We employ multiple prompt templates to avoid prompt-sensitive biases, and use their averaged performance as the final results. The best scores are in bold.}
\label{tab-proportion_result}
\end{table*}

\begin{table*}[t]
\small
\centering
\setlength{\tabcolsep}{3.5pt}
\renewcommand{\arraystretch}{1.1}
\begin{tabular}{lccccccccccccccccccc}
\hline
\multirow{2}{*}{} & & \multicolumn{2}{c}{Emotion}   &  & \multicolumn{2}{c}{Irony}      &  & \multicolumn{2}{c}{Stance}     &  & \multicolumn{2}{c}{MRPC}        &  & \multicolumn{2}{c}{RTE}         &  & \multicolumn{2}{c}{AVG}        \\ 
\cline{3-4} \cline{6-7} \cline{9-10} \cline{12-13} \cline{15-16} \cline{18-19} 
&  & $\sigma(\mathit{\delta_1})$        & $\sigma(\mathit{\delta_2})$        &  & $\sigma(\mathit{\delta_1})$         & $\sigma(\mathit{\delta_2})$        &  & $\sigma(\mathit{\delta_1})$         & $\sigma(\mathit{\delta_2})$        &  & $\sigma(\mathit{\delta_1})$         & $\sigma(\mathit{\delta_2})$         &  & $\sigma(\mathit{\delta_1})$         & $\sigma(\mathit{\delta_2})$         &  & $\sigma(\mathit{\delta_1})$         & $\sigma(\mathit{\delta_2})$        \\ \hline
\textit{orig} &                 & 3.85          & 1.69          &  & 4.37          & 2.71          &  & 4.85          & 1.03          &  & 5.85          & 1.52           &  & 4.11          & 1.49           &  & 4.61          & 1.69          \\
\textit{semt} &               & 1.25          & 0.69          &  & \cellcolor[HTML]{EFEFEF}\textbf{1.39}           & 1.79          &  & 3.00          & 1.14          &  & 2.77           & \cellcolor[HTML]{EFEFEF}\textbf{0.52}           &  & \cellcolor[HTML]{EFEFEF}\textbf{1.39}          & 0.75           &  & 1.96          & 0.98          \\
\textit{ours} &                   & \cellcolor[HTML]{EFEFEF}\textbf{0.60} & \cellcolor[HTML]{EFEFEF}\textbf{0.38} & \cellcolor[HTML]{EFEFEF} & 1.69  & \cellcolor[HTML]{EFEFEF}\textbf{1.32} & \cellcolor[HTML]{EFEFEF} & \cellcolor[HTML]{EFEFEF}\textbf{1.64}  & \cellcolor[HTML]{EFEFEF}\textbf{1.00} & \cellcolor[HTML]{EFEFEF} & \cellcolor[HTML]{EFEFEF}\textbf{0.63}  & 0.77  &  & 1.69  & \cellcolor[HTML]{EFEFEF}\textbf{0.44}  & \cellcolor[HTML]{EFEFEF} & \cellcolor[HTML]{EFEFEF}\textbf{1.25}  & \cellcolor[HTML]{EFEFEF}\textbf{0.78} \\ \hline
\end{tabular}
\caption{Standard deviations of data contamination resistance performance (\%) across different data proportions (20\%, 40\%, 60\%, 80\%, 100\%). The results are the mean values calculated across all eight open-source models. The best scores are in bold.}
\label{tab-proportion_std}
\end{table*}

\begin{table*}[]
\renewcommand{\arraystretch}{1.1}
\begin{tabular}{|ll|}
\hline
\multicolumn{2}{|c|}{\textbf{Guideline of Human Evaluation}}        \\ \hline

\multicolumn{2}{|c|}{\cellcolor[HTML]{EFEFEF}(1) Fluency} \\ \hline
\multicolumn{1}{|l|}{Definition} & \begin{tabular}[c]{@{}l@{}}Assess whether the language of the sentence is fluent, without grammatical or spelling \\errors. The scoring range is 1-3. \end{tabular} \\ \hline
\multicolumn{1}{|l|}{Score}    & \begin{tabular}[c]{@{}l@{}}1 point: The text contains multiple grammatical and/or spelling errors, significantly \\impacting the readability and understanding.\\2 points: The text contains a few grammatical or spelling errors, slightly affecting \\readability, but the overall meaning of the text is understandable.\\ 3 points: The text is grammatically and orthographically correct, expressing fluently \\and naturally, easy to understand.\end{tabular}    \\ \hline
\multicolumn{2}{|c|}{\cellcolor[HTML]{EFEFEF}(2) Coherence} \\ \hline
\multicolumn{1}{|l|}{Definition} &  \begin{tabular}[c]{@{}l@{}}Assess whether the question is logically clear and articulated explicitly. The scoring \\range is 1-3. \end{tabular}\\ \hline
\multicolumn{1}{|l|}{Score}    & \begin{tabular}[c]{@{}l@{}}1 point: The sentence lacks logical structure, is expressed in a disorganized manner, \\making it difficult for readers to understand. \\2 points: The sentence has a basic logical structure, with a relatively clear \\theme or argument, but the expression may not be \\direct enough or some parts may be slightly vague, affecting overall clarity. \\3 points: The question or answer has a clear structure, is logically coherent, expressed \\directly and clearly, easy to understand, and effectively conveys the theme or argument.\end{tabular}   \\ \hline
\multicolumn{2}{|c|}{\cellcolor[HTML]{EFEFEF}(3) Factuality}\\ \hline
\multicolumn{1}{|l|}{Definition} & \begin{tabular}[c]{@{}l@{}}Score based on whether [text] contains multiple factual errors, generally conforms to facts \\but contains minor errors or inaccuracies, or is entirely based on facts with all provided \\information being accurate. The scoring range is 0-1\end{tabular}\\ \hline
\multicolumn{1}{|l|}{Score}    & \begin{tabular}[c]{@{}l@{}}0 point: The text contains multiple factual errors or significant inaccuracies, making \\the information misleading or incorrect. \\ 1 point: The text is entirely factually accurate, with all provided information verified \\as correct.\end{tabular}\\ \hline
\multicolumn{2}{|c|}{\cellcolor[HTML]{EFEFEF}(4) Accuracy}\\ \hline
\multicolumn{1}{|l|}{Definition} & \begin{tabular}[c]{@{}l@{}}Score the category accuracy considering if the label category matches the content, \\ranging from 0-1.\end{tabular}\\ \hline
\multicolumn{1}{|l|}{Score}    & \begin{tabular}[c]{@{}l@{}}0 point: The assigned label does not match the content or is misleading. \\ 1 point: The assigned label accurately reflects the content.\end{tabular}\\ \hline

\end{tabular}
\caption{Guideline of human evaluation for data quality.}
\label{tab-human}
\end{table*}

\end{document}